\pdfoutput=1
\documentclass[11pt]{article}

\usepackage[preprint]{acl}

\usepackage{times}
\usepackage{latexsym}

\usepackage[T1]{fontenc}
\usepackage[utf8]{inputenc}

\usepackage{microtype}

\usepackage{inconsolata}

\usepackage{graphicx}

\usepackage{enumitem}
\usepackage{etoolbox}
\usepackage{booktabs,makecell,tabularx} 
\usepackage{resizegather}
\usepackage{multirow}
\usepackage{amsmath,amsfonts,amssymb} % 需加载必要的数学宏包
\usepackage{graphicx}
\usepackage{tcolorbox}
\usepackage{subcaption}
\usepackage{colortbl}
\usepackage{listings}
\usepackage{makecell}
\usepackage{multirow}
\usepackage{xcolor}
\usepackage{subcaption} % 用于插入子图
\usepackage{tcolorbox}
\usepackage{longtable}
\tcbuselibrary{listings,skins}
\usepackage{geometry}  % 可选，调整页边距，确保表格不会超出页面
\usepackage{xspace}

\definecolor{codebg}{rgb}{0.95,0.95,0.95}
\definecolor{codeframe}{rgb}{0.5,0.5,0.5}

\tcbuselibrary{listings,skins}

\newtcblisting{mycode}{
    listing engine=listings,
    colback=white,
    colframe=gray!75!black,
    listing only,
    left=1mm,
    enhanced,
    overlay={
        \begin{tcbclipinterior}
            \fill[gray!25] (frame.south west) rectangle ([xshift=1mm]frame.north west);
        \end{tcbclipinterior}
    },
    boxrule=0.5pt
}

\newcommand{\M}{Character-R1\xspace}

\definecolor{deepgreen}{rgb}{0.0, 0.6, 0.0}

\title{Character-R1: Enhancing Role-Aware Reasoning \\ in Role-Playing Agents via RLVR}

\author{Yihong Tang$^{1,2}$, Kehai Chen$^{1,}$\thanks{Corresponding author.}, Xuefeng Bai$^{1}$, \\ {\bf Benyou Wang$^{3,2}$, Zeming Liu$^4$, Haifeng Wang$^5$, Min Zhang$^1$} \\
        $^1$Institute of Computing and Intelligence, Harbin Institute of Technology, Shenzhen, China \\
        $^2$Shenzhen Loop Area Institute (SLAI), Shenzhen, China \\
        $^3$The Chinese University of Hong Kong, Shenzhen, China \\
        $^4$Beijing University of Aeronautics and Astronautics, Beijing, China \\
        $^5$Baidu Inc., Beijing, China \\
        \texttt{\{toyhom@stu.hit.edu.cn,  chenkehai@hit.edu.cn\}}
}

\begin{document}
\maketitle
\begin{abstract}
% Current role-playing agents (RPAs) are typically constructed by imitating surface-level behaviors, but lack internal cognitive consistency, often causing out-of-character errors in complex situations. 
% Although recent studies attempt to introduce reasoning capabilities into these agents, existing methods still fail to provide direct and comprehensive reward signals to effectively guide reasoning. 

Current role-playing agents (RPAs) are typically constructed by imitating surface-level behaviors, but this approach lacks internal cognitive consistency, often causing out-of-character errors in complex situations.
To address this, we propose Character-R1, a framework designed to provide comprehensive verifiable reward signals for effective role-aware reasoning, which are missing in recent studies.
Specifically, our framework comprises three core designs: (1) Cognitive Focus Reward, which enforces explicit label-based analysis of 10 character elements (e.g., worldview) to structure internal cognition; (2) Reference-Guided Reward, which utilizes overlap-based metrics with reference responses as optimization anchors to enhance exploration and performance; and (3) Character-Conditioned Reward Normalization, which adjusts reward distributions based on character categories to ensure robust optimization across heterogeneous roles. 
Extensive experiments demonstrate that Character-R1 significantly outperforms existing methods in knowledge, memory and others. \footnote{https://github.com/Toyhom/Character-R1.}
\end{abstract}

\section{Introduction}

\begin{figure}
    \centering
    \includegraphics[width=0.98\linewidth]{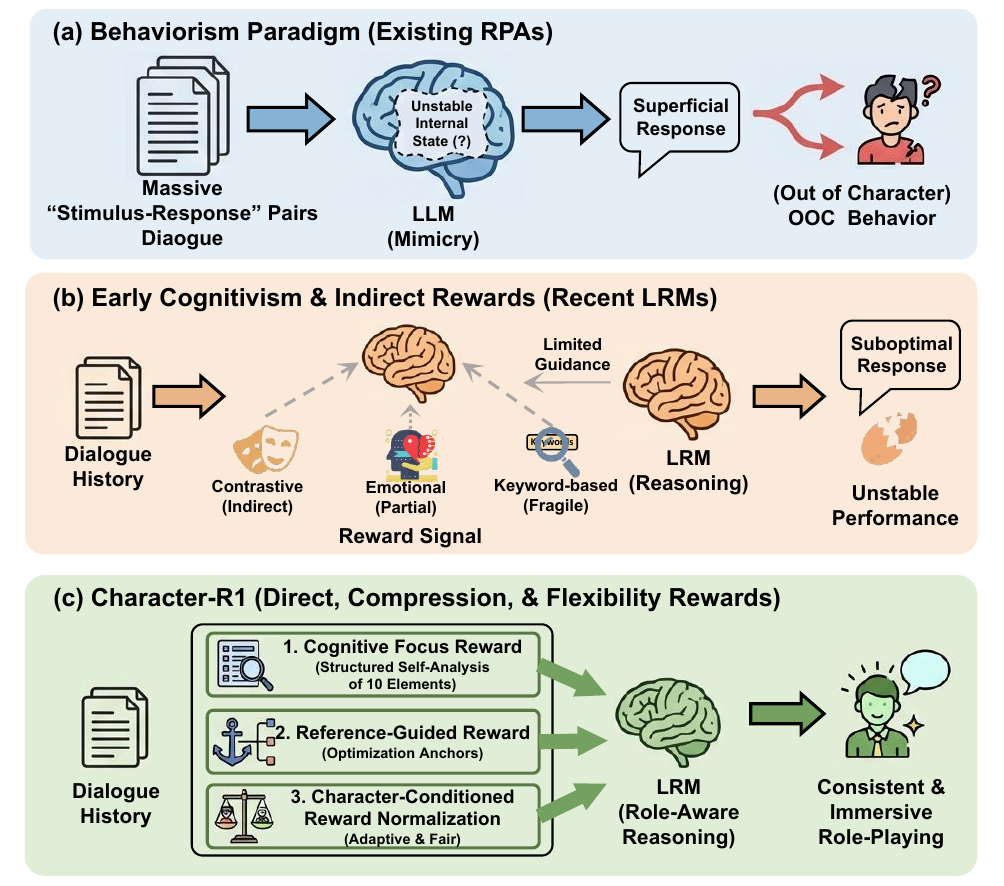}
    \caption{Comparison of role-playing paradigms. (a) Behaviorism mimics surface patterns,leading to OOC behavior; (b) Early cognitivism uses indirect,partial,and fragile rewards;(c) \M employs direct,comprehensive,and flexible rewards to explicitly shape the character's internal cognitive state.}
    \label{fig:motivation}
    \vspace{-0.6cm}
\end{figure}

With the development of Large Language Models (LLMs), Role-Playing Agents (RPAs)~\citep{chen2024oscarsaitheatersurvey} have attracted widespread attention in applications such as emotional companionship~\citep{liu2024speak}, virtual interaction~\citep{park2023generative}, and social simulation~\citep{wang2025yulan}. 
Most existing works follow the \textit{behaviorism paradigm} RPAs by fine-tuning on dialogue datasets~\citep{zhou2024characterglm} or enhancing character trait utilization~\citep{yu2024neeko}.
As illustrated in Figure \ref{fig:motivation}(a), this paradigm learns by imitating massive ``stimulus-response'' pairs (i.e., dialogue turns). It focuses solely on generating superficially plausible responses while neglecting the internal cognitive consistency of the character.
Consequently, when facing complex or unseen scenarios, these agents frequently exhibit out-of-character (OOC) behavior.
% Consequently, when facing complex, contradictory, or unseen dialogue situations and characters, these agents frequently exhibit Out-of-Character (OOC).

This issue prompted recent research to introduce Large Reasoning Models (LRMs)~\citep{guo2025deepseek} from a \textit{cognitivism} perspective to explicitly model the internal mental state or reasoning process, termed role-aware reasoning~\citep{tang2025thinking}. 
However, direct application of general LRMs compromises role-playing performance \cite{feng2025reasoning}, as they are typically trained on scenarios like coding and mathematics that differ significantly from role-playing. Therefore, explicit character-specific reward signals are necessary to guide Reinforcement Learning (RL) training.
\citet{tang2025thinking} adopt contrastive learning based on self-awareness and scene differences as an indirect signal; \citet{wang2025rlver} design a verifiable emotional reward, utilizing sentiment classifiers to calibrate the model's expression; and \citet{wang2025raidenr1} employ a keyword-matching-based verifiable reward. Despite their success, these approaches represent limited reward guidance, as their signals tend to be indirect, partial, and fragile (see Figure \ref{fig:motivation}(b)).
% While existing works utilize contrastive learning \cite{tang2025thinking} or verifiable rewards based on sentiment and keywords \cite{wang2025rlver, wang2025raidenr1}, these signals remain indirect and fragile (Figure \ref{fig:motivation}(b)). Such partial rewards fail to comprehensively guide the model in maintaining consistent character cognition.

To address these limitations, we propose \M, a role-aware reasoning framework driven by direct, comprehensive, and flexible reward signals (Figure \ref{fig:motivation}(c)). \M comprises three core designs: (1) Cognitive Focus Reward: Requires the model to construct the character's reasoning process through explicit label-based analysis of 10 core elements (e.g., knowledge background, behavioral style, worldview), thereby locking onto character cognition; (2) Reference-Guided Reward: Leverages reference responses as flexible optimization anchors to facilitate rapid exploration convergence, while surpassing the reference; (3) Character-Conditioned Reward Normalization: Addresses the heterogeneity of reward distributions across different characters by normalizing relative to character categories and reward types globally, ensuring the model efficiently extracts optimization signals from multi-dimensional rewards.

Experimental results on role-playing benchmarks indicate that \M significantly outperforms existing methods, especially in knowledge, persona and memory, demonstrating the effectiveness of our proposed multi-dimensional, verifiable reward framework. The main contributions of this study are as follows:
\begin{itemize}
\item We design a set of direct, broad, and flexible RL reward mechanisms for role-aware reasoning, addressing the challenge that rewards in role-playing agents are difficult to verify objectively and are insufficiently comprehensive.
\item We propose Character-Conditioned Reward Normalization to mitigate reward heterogeneity arising from diverse character traits, ensuring fair optimization across all roles.
\item We demonstrate through extensive experiments that \M significantly enhances the character consistency and immersive capabilities of role-playing agents.
\end{itemize}

\section{Related Work}
\paragraph{Role-Playing Agents with LLMs}
The rapid development of Large Language Models (LLMs) and the substantial demand for emotional companionship lay the foundation for research on Role-Playing Agents (RPAs)~\citep{chen2024oscarsaitheatersurvey}. Early research primarily relies on the In-Context Learning capabilities of LLMs or utilizes synthetic high-quality dialogues~\citep{tang-etal-2023-enhancing-personalized, Wei2023MultiPartyCC, Mao2023EditingPF, Wang2023DoesRC, wang-etal-2024-incharacter} and corpora extracted from scripts~\citep{shao2023character}, novels~\citep{Xu2024CharacterID}, and real-person role-playing logs~\citep{zhou2024characterglm}. Furthermore, some studies are dedicated to enhancing the model's ability to model specific character traits, like Neeko~\citep{yu2024neeko}, HIRPF~\citep{Sun2024IdentityDrivenHR} and RoleMRC \citep{lu2025rolemrc}. More recently, the field has shifted towards cognitive simulation by explicitly modeling the character's thought process~\citep{wang2025coser}. These works have introduced supervision signals for this purpose, such as contrastive learning for reasoning styles \cite{tang2025thinking} or rewards based on sentiment and keywords \cite{wang2025rlver, wang2025raidenr1}.

% Role-playing dialogue agents~\citep{chen2024oscarsaitheatersurvey} have emerged as a flourishing research field alongside the advancement of Large Language Models (LLMs). Early approaches~\citep{tang-etal-2023-enhancing-personalized, Wei2023MultiPartyCC, Mao2023EditingPF, Wang2023DoesRC, wang-etal-2024-incharacter} primarily rely on LLMs' in-context learning (ICL)~\citep{dong2024surveyincontext,zhu2024towardsrobust} capabilities. Subsequent research recognizes the importance of specialized role-playing models, leading to efforts in synthesizing data at scale using stronger models~\citep{wang2024rolellm} or extracting conversations from scripts~\citep{shao2023character}, novels~\citep{Xu2024CharacterID}, and live role-playing sessions~\citep{zhou2024characterglm}. 

% Concurrently, recent works utilize explicit modeling of the character's thought process. Tang et al. (2025) propose the RAR framework, which utilizes contrastive learning to indirectly shape reasoning styles. Wang et al. (2025) introduce verifiable rewards based on sentiment classifiers, focusing supervision signals on the emotional dimension. Going further, Wang et al. (2025) use keyword-based rewards in RAIDEN-R1 to directly quantify the application of character knowledge. These works promote a paradigm shift in RPAs from "behavioral imitation" to "cognitive simulation" by introducing supervision signals that incentivize the model to "think like a character." 

However, these studies either focus heavily on superficial imitation of character behavior while neglecting the modeling of internal cognitive processes, or they focus only on a single dimension of the thinking process, causing the model to struggle in maintaining consistent logic during complex situations. Unlike these works, \M realizes comprehensive, direct, and flexible supervision of the character's cognitive process, precisely addressing the deficiencies in supervision signals found in existing methods.

\begin{figure*}[htbp]
\centering
\includegraphics[width=0.97\textwidth]{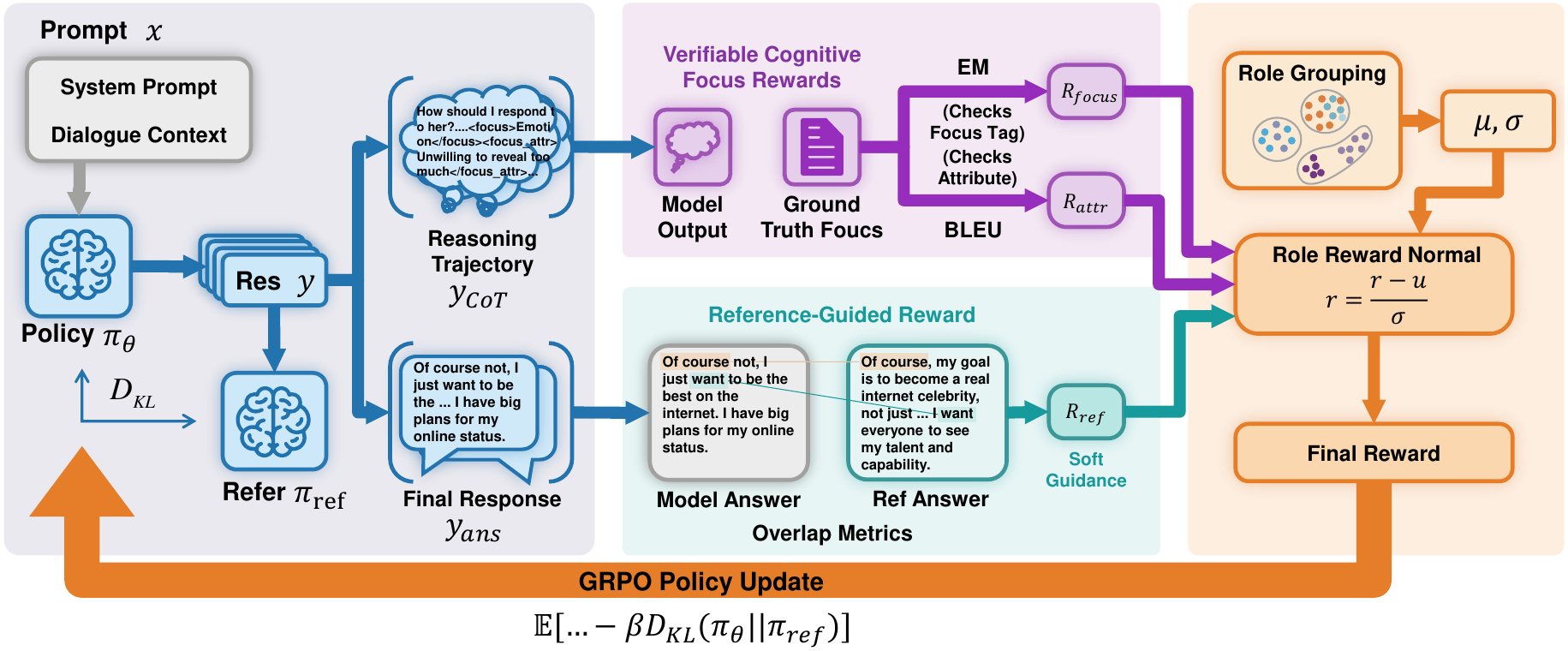}
\caption{Overview of the \M methodology. The layout illustrates the generation of the reasoning trajectory ($y_{CoT}$) containing explicit verifiable <focus> tags, followed by the response ($y_{ans}$). Rewards are derived from three sources: Cognitive Focus accuracy, Focus Attribute quality, and Reference Guidance. These rewards undergo Character-Conditional Normalization utilizing Role Grouping statistics before being aggregated to update the policy via GRPO.}
\label{fig: main}
% \vspace{-0.6cm}
\end{figure*}

\paragraph{Large Reasoning Models}

Large Reasoning Models (LRMs)~\citep{tie2025survey,kumar2025llm,li2025system,xu2025towards,chen2025towards,bandyopadhyay2025thinking} have demonstrated remarkable success in decomposing complex problems into interpretable chains of thought. Existing research mainly employs three strategies to cultivate the deep reasoning abilities of LRMs: first, Knowledge Distillation, which efficiently transfers structured reasoning patterns from large teacher models to smaller student models~\citep{muennighoff2025s1,ye2025limo,li2025small,li2025llms,openthoughts}; second, auxiliary enhancement techniques, such as combining process supervision~\citep{Lightman2023LetsVS, zhu2024benchmarking} or advanced search algorithms~\citep{wan2024alphazero, xu2025memory, xin2024deepseekprover} to improve reasoning quality; and most significantly, large-scale Reinforcement Learning (RL) training. This approach serves as the core engine for optimizing the model's intrinsic thinking and planning capabilities, leveraging carefully designed reward signals and massive computing power to achieve superior performance~\citep{jaech2024openai,deepmind2025blog,guo2025deepseek,team2025kimi,qwen2025qwqblog,openai2025o3blog}.

However, the aforementioned methods are designed for logical and factual tasks (e.g., math, code), where correctness can be verified by objective standards. Distinct from them, our research focuses on designing a comprehensive, verifiable reward tailored specifically for role-playing scenarios, and developing corresponding training techniques.

\section{Method}

In this section, we present the \M, a framework designed to enhance character role-playing through reinforcement learning with verifiable rewards. We begin by reviewing the preliminary concepts of Group Relative Policy Optimization (GRPO)~\citep{guo2025deepseek}. We then introduce our novel reward mechanisms: the \textit{Cognitive Focus Task}, which structures the model's reasoning process using verifiable internal states, and \textit{Reference Guidance}, which utilizes overlap-based metrics to close to the reference. Finally, we describe our \textit{Character-Conditional Reward Normalization}, a component that stabilizes training by addressing distributional shifts across different characters and reward types. Figure~\ref{fig: main} provides an overview of the proposed framework.

\subsection{Preliminaries}
GRPO is a variant of Proximal Policy Optimization (PPO)~\citep{schulman2017ppo} that eliminates the value function. It estimates the advantage function by normalizing rewards across multiple sampled outputs within a batch. For a specific data pair $(x, y)$, the behavioral policy $\pi_{\theta_{old}}$ samples a set containing $G$ independent responses $\{\hat{y_i}\}_{i=1}^{G}$. Each response $\hat{y}_i$ receives a scalar reward $R_i$, and its advantage function $\hat{A}_i$ is calculated via group-wise normalization: $\hat{A}_{i,t} = \frac{R_i - \text{mean}\left(\{R_i\}_{i=1}^{G}\right)}{\text{std}\left(\{R_i\}_{i=1}^{G}\right)+\epsilon}$.
% \begin{equation}
% \small
% \begin{aligned}
% \hat{A}_{i,t} = \frac{R_i - \text{mean}\left(\{R_i\}_{i=1}^{G}\right)}{\text{std}\left(\{R_i\}_{i=1}^{G}\right)+\epsilon}.
% \end{aligned}
% \end{equation}

The objective of GRPO combines the clipped objective of PPO with a KL divergence penalty term to achieve stable and efficient policy updates:
\begin{equation}
\small
\begin{aligned}
    \mathcal{J}_{GRPO}(\theta) =& \mathbb{E}_{(x,y)\sim \mathcal{D},\{\hat{y_i}\}_{i=1}^{G}\sim \pi_{\theta_{old}}(\cdot|x)} \bigg[ \frac{1}{G}\sum_{i=1}^{G} \frac{1}{|\hat{y_i}|}\sum_{t=1}^{|\hat{y_i}|} \\
    &\bigg( \min\bigl( r_{i,t}(\theta)\hat{A}_{i,t}, \text{clip}\bigl(r_{i,t}(\theta), 1-\varepsilon, 1+\varepsilon\bigr)\hat{A}_{i,t} \bigr) 
    \\
    &- \beta D_{KL}\bigl(\pi_{\theta} \parallel \pi_{ref}\bigr) \bigg) \bigg],
\end{aligned}
\end{equation}
where:
$r_{i,t}(\theta) = \frac{\pi_{\theta}(\hat{y}_{i,t} \mid x, \hat{y}_{i,<t})}{\pi_{\theta_{old}}(\hat{y}_{i,t} \mid x, \hat{y}_{i,<t})}$,
$\mathcal{J}_{GRPO}(\theta)$ is the objective function of GRPO, and $|\hat{y_i}|$ denotes the token length of the $i$-th response. $\hat{y}_{i,t}$ represents the $t$-th token in the $i$-th response, while $\hat{y}_{i,<t}$ denotes the sequence of tokens preceding the $t$-th token. $\beta$ is the coefficient for the KL penalty term, $\pi_{ref}$ is the reference model, and $D_{KL}(\pi_{\theta} \parallel \pi_{ref})$ represents the KL divergence between the $\pi_{\theta}$ and the $\pi_{ref}$.

In traditional LLM reinforcement learning, reward signals typically originate from an uninterpretable reward model, which may lead to reward hacking. To address this, many works turn to verifiable rewards, implying that reward signals stem from programmatically verifiable objective facts. The model is trained to generate a sequence containing a reasoning trajectory $y_{CoT}$ and a final answer $y_{ans}$:
$y = y_{CoT} \oplus y_{ans}$,
where $\oplus$ denotes concatenation. Role-aware reasoning requires the model to incorporate the character's mindset during the reasoning process $y_{CoT}$ and think like the character. However, in open-ended tasks such as role-playing, there is no objective final answer, making it difficult to incentivize the model to reason and explore during the thinking process.

\subsection{Cognitive Focus}
To introduce verifiable rewards into role-playing, we design the Cognitive Focus Task. This task mandates that the model explicitly declare and elaborate on its current cognitive focus within its reasoning trajectory $y_{CoT}$ before generating a response. Cognitive foci comprise 10 core cognitive dimensions, including knowledge, style, worldview, and emotion, etc. The model is required to spontaneously determine which dimensions to focus on in the current context and provide specific attribute descriptions. This process is verifiable, corresponding to two rewards that quantify the model's performance on the cognitive focus task:
\begin{equation}
\small
\begin{aligned}
R_{focus}=EM(y_{focus}^{pred},y_{focus}^{gold}), \\
R_{focus\_attr}=BLEU(y_{focus\_attr}^{pred},y_{focus\_attr}^{gold}).
\end{aligned}
\end{equation}

Where $y_{focus}$ and $y_{focus\_attr}$ represent the cognitive focus and its specific attributes, respectively, which are included in $y_{CoT}$. $R_{focus}$ is the focus accuracy reward, assessing whether the model selects the correct cognitive dimension and is measured by the Exact Match (EM). $R_{focus\_attr}$ is the focus attribute reward, evaluating the accuracy of the description of the selected focus, measured by calculating the BLEU-1. The cognitive focus rewards compel the model to activate character in a multi-perspective and structured manner, forming a coherent and interpretable internal state.  Details of the method can be found in Appendix~\ref{app: method}.

\subsection{Reference Guidance}
In the response generation phase, we design the reference-guided reward to impose certain constraints on the response. Specifically, we ingeniously employ overlap-based metrics (such as BLEU~\citep{papineni2002bleu}), which are traditionally used in text generation tasks, as a form of soft guidance reward:
$
R_{ref}=\sum{OverlapMetric(y_{ans}^{pred},y_{ans}^{refer})},
$
where $OverlapMetric$ denotes an overlap-based metric. This design functions as a soft constraint mechanism. In the initial stage of model optimization, when the generated responses differ significantly from the reference responses in style, $R_{ref}$ provides dense reward signals to rapidly guide the model toward convergence in the right direction. 

As the generation quality and similarity to the reference response improve, the exploration sampling mechanism of RL ensures that the model does not strictly generate the exact reference response. Consequently, this forms a soft constraint that prevents the model from overfitting to the reference. Ultimately, the reference guidance reward points out the general direction for optimization while preserving space for exploring better solutions.

\subsection{Character-Conditional Reward Normalization}
Through the aforementioned designs, we obtain a multi-dimensional reward vector $\mathbf{r} \in \mathbb{R}^K$ for each generated response $y$. However, due to substantial differences in reward distributions across different characters and reward categories, direct weighted summation leads to certain characters or reward types dominating the gradient updates inappropriately.

To address this issue, we propose Character-Conditional Reward Normalization, which performs normalization at two levels—character and reward type. Specifically, we maintain an independent global normalization instance for each character $c$ and each reward type $k$:
$ \hat{r}_{c,k} = \frac{r_k - \mu_{c,k}}{\sqrt{\sigma^2_{c,k} + \epsilon}}$, 
where $\mu_{c,k}$ and $\sigma^2_{c,k}$ are the running mean and variance specific to character $c$ and reward $k$, updated dynamically during training via exponential moving average.

\begin{table}[h]
\centering
% \Large  <-- 建议去掉这个，通常不需要在 resizebox 内部强行放大字体
\resizebox{.48\textwidth}{!}{ % .99 有时会太满，.95 留一点白边更美观
\begin{tabular}{lcccc}
\toprule
  Model & Backbone & Data Scale & Method \\ % 
\midrule
    CharacterGLM-6B~\citep{zhou2024characterglm} & ChatGLM2-6B & 1k & SFT \\
    Peach-9B-8k-Roleplay \citep{closedcharacter2024peach9b} & Yi-1.5-9B & 100k & SFT \\
    Llama-3.1-8B-RoleMRC \citep{lu2025rolemrc} & Llama-3.1-8B & 107.7k & SFT, DPO \\
    Haruhi-Zero-7B \citep{silkroad2024haruhizero} & Qwen-7B & 120k & SFT \\
    Crab \citep{he2025crab} & Llama-3.1-8B & 41k & SFT \\
    CoSER-Llama-3.1-8B \cite{wang2025coser} & Llama-3.1-8B & 29.7k & SFT \\
    Peach-2.0-9B-8k-Roleplay \cite{closedcharacter2024peach2} & Yi-1.5-9B & 100k & SFT, DPO \\
    Qwen2.5-7B-RoleMRC \citep{lu2025rolemrc} & Qwen2.5-7B & 107.7k & SFT, DPO \\ 
    Humanish-Roleplay \cite{vicgalle2024humanish} & Llama-3.1-8B & 269.9k & SFT, DPO 
    \\
    \midrule
    Ours & $-$ & 650 & RL \\
\bottomrule
\end{tabular}}
\caption{Details of the specialized role-playing models.} % 更专业的 caption
\label{tab: model_specs} % 更准确的 label
\end{table}

\begin{table*}[t!]
\centering
\Large
\resizebox{.99\textwidth}{!}{
\begin{tabular}{lcccccccccccccc}
\toprule
\multicolumn{15}{l}{$MC$: Memory Consistency\quad $FA$: Fact Accuracy\quad $BC_K$: Boundary Consistency\quad $AC^b$: Attribute Consistency(Bot)}
\\
\multicolumn{15}{l}{$AC^h$: Attribute Consistency(Human)\quad $BC^b_P$: Behavior Consistency(Bot)\quad $BC^h_P$: Behavior Consistency(Human)\quad $HL$:Human-likeness} \\

\multicolumn{15}{l}{$ES$: Emotional Self-regulation\quad $ER$: Empathetic Responsiveness\quad $MS$: Morality Stability $MR$: Morality Robustness\quad $EG$:Engagement} \\
\midrule
  \multirow{2}{*}{\textbf{Method}} & \multicolumn{1}{c}{\textbf{Memory}} & \multicolumn{2}{c}{\textbf{Knowledge}} & \multicolumn{4}{c}{\textbf{Persona}} & \multicolumn{2}{c}{\textbf{Emotion}} & \multicolumn{2}{c}{\textbf{Morality}} & \multicolumn{2}{c}{\textbf{ Believability}} & \multirow{2}{*}{\textbf{Avg.}}\\
\cmidrule(lr){2-2}\cmidrule(lr){3-4}\cmidrule(lr){5-8}\cmidrule(lr){9-10}\cmidrule(lr){11-12}\cmidrule(lr){13-14}
  & $MC$ & $FA$ & $BC_K$ & $AC^b$ & $AC^h$ & $BC^b_P$ & $BC^h_P$ & $ES$ & $ER$ & $MS$ & $MR$ & $HL$ & $EG$ & \\
\midrule
        Qwen2.5-7B-Instruct & 3.938  & 2.356  & 3.800  & 4.631  & 4.275  & 3.856  & 3.233  & 3.238  & 3.097  & 4.918  & 4.861  & \underline{3.120}  & 3.120  & 3.726  \\ 
        \quad\textit{+} SFT & 4.020  & 2.165  & 3.873  & 4.240  & 4.105  & 3.545  & 3.207  & 3.170  & 2.771  & 4.720  & 4.788  & 3.000  & 3.056  & 3.589  \\ 
        \quad\textit{+} Neeko & 3.997  & 2.192  & 3.775  & 4.257  & 3.288  & 3.212  & 3.084  & 3.229  & 2.693  & 4.550  & 4.703  & 2.799  & 2.942  & 3.440  \\ 
        \quad\textit{+} RAR & 4.059  & \underline{2.585}  & 3.934  & 4.314  & 4.272  & \textbf{4.090}  & \textbf{3.977}  & 3.145  & 2.809  & 4.825  & 4.808  & 2.814  & 2.946  & 3.737  \\ 
        \quad\textit{+} \textbf{\M} & \textbf{4.444}  & \textbf{2.688}  & \textbf{4.017}  & 4.544  & 4.381  & 3.956  & 3.317  & \textbf{3.463}  & \underline{3.072}  & \underline{4.925}  & \underline{4.861}  & 3.030  & 3.115  & \textbf{3.832}  \\ 
        \midrule 
        Llama-3.2-3B-Instruct & 3.588  & 2.088  & 3.892  & 4.606  & 4.300  & 4.025  & 3.300  & 3.306  & 3.053  & 4.840  & 4.649  & 2.890  & 3.320  & 3.681  \\ 
        \quad\textit{+} SFT & 3.606  & 2.063  & 3.892  & 4.550  & 4.169  & 3.894  & 3.500  & 3.219  & 3.003  & 4.888  & 4.823  & 3.350  & 3.310  & 3.713  \\ 
        \quad\textit{+} Neeko & 3.796  & 1.855  & 3.816  & 4.513  & 3.428  & 3.416  & 3.192  & 3.319  & 2.747  & 4.561  & 4.582  & 2.665  & 2.972  & 3.451  \\ 
        \quad\textit{+} RAR & 3.848  & 2.010  & 3.978  & 4.511  & 4.373  & 3.918  & 4.112  & 3.240  & 2.910  & 4.785  & 4.743  & 2.718  & 2.964  & 3.701  \\ 
        \quad\textit{+} \textbf{\M} & \underline{4.275}  & 2.144  & 3.983  & \textbf{4.950}  & \textbf{4.556}  & \underline{4.038}  & 3.350  & \underline{3.463}  & \textbf{3.134}  & 4.850  & 4.748  & 2.940  & 2.970  & \underline{3.800}  \\ 
        \midrule  
        CharacterGLM-6B & 3.245  & 2.100  & 3.543  & 3.365  & 3.410  & 3.070  & 3.100  & 2.610  & 2.500  & 4.480  & 4.800  & 2.840  & 2.700  & 3.213  \\ 
        Peach-9B-8k-Roleplay & 2.931  & 2.344  & 3.717  & 3.800  & 3.388  & 3.300  & 3.292  & 2.863  & 2.802  & 4.838  & 4.798  & 2.475  & 2.435  & 3.306  \\ 
        Llama-3.1-8B-RoleMRC & 4.169  & 2.163  & 3.717  & 4.588  & 4.231  & 3.606  & 3.225  & 3.125  & 2.864  & 4.800  & 4.748  & 3.040  & 3.060  & 3.641  \\ 
        Haruhi-Zero-7B & 4.088  & 2.150  & 3.583  & 4.525  & 4.256  & 3.613  & 3.192  & 3.456  & 2.965  & 4.850  & 4.773  & 2.990  & 3.065  & 3.654  \\ 
        Crab & 4.131  & 2.175  & 3.600  & 4.650  & 4.250  & 3.681  & 3.167  & 3.331  & 2.965  & 4.850  & 4.684  & 3.050  & 3.110  & 3.665  \\ 
        CoSER-Llama-3.1-8B & 4.056  & 2.113  & 3.958  & 4.506  & 4.213  & 3.700  & 3.250  & 3.156  & 2.883  & \textbf{4.950}  & \textbf{4.874}  & 3.045  & 2.970  & 3.667  \\ 
        Peach-2.0-9B-8k-Roleplay & 2.819  & 2.400  & 3.425  & 4.400  & 4.263  & 4.088  & \underline{3.867}  & 3.350  & 3.028  & 4.475  & 4.470  & \textbf{3.485}  & \textbf{3.680}  & 3.673  \\ 
        Qwen2.5-7B-RoleMRC & 4.175  & 2.231  & 3.775  & \underline{4.713}  & 4.300  & 3.688  & 3.225  & 3.188  & 2.927  & 4.750  & 4.735  & 3.000  & 3.105  & 3.678  \\ 
        Humanish-Roleplay & 4.013  & 2.350  & \underline{4.000}  & 4.694  & \underline{4.481}  & 3.975  & 3.267  & 3.444  & 3.053  & 4.750  & 4.823  & 3.010  & \underline{3.145}  & 3.770  \\ 
        \midrule 
        CharacterGLM & 3.760  & 2.180  & 3.970  & 4.030  & 3.800  & 3.260  & 2.890  & 2.940  & 2.640  & 4.530  & 4.510  & 3.160  & 3.320  & 3.461  \\ 
        Llama-3-70B-Instruct & 3.940  & 2.590  & 3.950  & 4.390  & 3.960  & 3.330  & 3.350  & 3.060  & 2.890  & 4.710  & 4.740  & 3.400  & 3.510  & 3.678  \\ 
        GLM-4 & 3.524  & 2.373  & 3.701  & 4.380  & 4.103  & 3.728  & 3.487  & 3.130  & 2.987  & 4.826  & 4.876  & 3.208  & 3.500  & 3.679  \\ 
        Baichuan-NPC & 3.672  & 2.134  & 4.132  & 4.254  & 4.216  & 4.022  & 3.366  & 3.001  & 3.179  & 4.830  & 4.897  & 2.975  & 3.297  & 3.690  \\ 
        GPT-3.5-turbo & 3.490  & 2.451  & 3.692  & 4.345  & 4.155  & 3.635  & 3.560  & 3.090  & 2.838  & 4.735  & 4.761  & 3.619  & 3.758  & 3.702  \\ 
        GPT-4o & 3.793  & 2.647  & 3.978  & 4.484  & 4.027  & 3.723  & 3.414  & 3.046  & 2.974  & 4.763  & 4.771  & 3.261  & 3.510  & 3.722  \\ 
        Qwen2.5-72B-Instruct & 4.060  & 2.558  & 4.102  & 4.531  & 3.222  & 3.917  & 3.439  & 3.398  & 2.962  & 4.897  & 4.815  & 3.512  & 3.418  & 3.756  \\ 
        Gemini-3-pro & 3.946  & 2.905  & 4.099  & 4.167  & 4.133  & 3.671  & 3.529  & 3.319  & 2.973  & 4.820  & 4.633  & 3.613  & 3.423  & 3.787  \\ 
        Claude-4-opus & 3.997  & 2.435  & 4.398  & 4.508  & 4.362  & 3.877  & 3.751  & 3.587  & 3.176  & 4.936  & 4.735  & 3.151  & 3.351  & 3.866 \\ 

\bottomrule
\end{tabular}}
\caption{Performance comparison of different methods on the CharacterBench. The best value (excluding large foundation models) for each metric is in \textbf{bold}, and the second-best value is \underline{underlined}.}
\label{tab: main_characterbench}
\end{table*}

Furthermore, to address sparsity from excessive roles, we introduce Role Grouping. Character descriptions are encoded via an embedding model and partitioned into groups using k-means~\citep{lloyd1982least}.
Subsequently, normalization is performed at the granularity of Role Groups, thereby sharing statistical information among similar characters and enhancing the robustness of normalization.

Finally, we compute the dot product of the normalized reward vector $\hat{\mathbf{r}}$ and a predefined weight vector $\mathbf{w} \in \mathbb{R}^K$ to obtain the final scalar reward:
$R = \mathbf{w}^T \hat{\mathbf{r}} = \sum_{j=1}^{K} w_j \hat{r}_j$. 
Through this series of designs, we successfully extract a verifiable reward signal that can be effectively utilised in open-ended role-playing tasks.

\section{Experiments}

\begin{table*}[t!]
\centering
\scriptsize
\resizebox{.99\textwidth}{!}{
\begin{tabular}{lcccccccccc}
\toprule

\multicolumn{11}{l}{Sty.: Role Style\quad Konw.: Role Knowledge\quad SU: Situational Understanding\quad ED: Emotion Detection\quad HSD: Humor Sarcasm Detect}
\\
\multicolumn{11}{l}{MEM: Long-Term Conversation Memory\quad Neu., Pos., Neg.: Social Preference}
\\
\midrule

   \textbf{Method}  & \textbf{Know.} & \textbf{Sty.} & \textbf{ED} & \textbf{SU} & \textbf{HSD} & \textbf{MEM}  & \textbf{Neu.} & \textbf{Pos.} & \textbf{Neg.} & \textbf{Avg.} \\
\midrule
        Qwen2.5-7B-Instruct & \underline{0.904}  & 0.793  & 0.341  & 0.308  & 0.870  & 0.752  & 0.869  & 0.903  & 0.802  & 0.727  \\
        \quad\textit{+} SFT & 0.903  & 0.777  & 0.394  & 0.333  & 0.900  & 0.760  & \underline{0.872}  & 0.911  & 0.824  & 0.742  \\
        \quad\textit{+} Neeko & \textbf{0.917}  & 0.738  & 0.338  & 0.383  & 0.869  & 0.726  & 0.829  & 0.901  & 0.798  & 0.722  \\
        \quad\textit{+} RAR & 0.881  & 0.728  & 0.400  & 0.339  & \textbf{0.937}  & \textbf{0.877}  & 0.861  & \underline{0.904}  & \underline{0.835}  & \underline{0.751}  \\
        \quad\textit{+} \M & 0.896  & \underline{0.795}  & 0.408  & 0.433  & \underline{0.930}  & 0.841  & \textbf{0.875}  & \textbf{0.920}  & 0.824  & \textbf{0.769}  \\
        \midrule
        Llama-3.2-3B-Instruct & 0.735  & 0.725  & 0.404  & 0.354  & 0.730  & 0.504  & 0.769  & 0.805  & 0.669  & 0.633  \\
        \quad\textit{+} SFT & 0.747  & 0.735  & 0.400  & 0.304  & 0.730  & 0.494  & 0.769  & 0.835  & 0.662  & 0.631  \\
        \quad\textit{+} Neeko & 0.771  & 0.650  & 0.398  & 0.296  & 0.576  & 0.446  & 0.728  & 0.816  & 0.600  & 0.587  \\
        \quad\textit{+} RAR & 0.781  & 0.705  & 0.415  & 0.296  & 0.735  & 0.512  & 0.737  & 0.797  & 0.635  & 0.624  \\
        \quad\textit{+} \M & 0.847  & 0.727  & 0.405  & 0.267  & 0.730  & 0.564  & 0.776  & 0.816  & 0.676  & 0.645  \\
        \midrule
        Peach-9B-8k-Roleplay & 0.580  & 0.253  & 0.408  & 0.313  & 0.670  & 0.464  & 0.609  & 0.671  & 0.484  & 0.494  \\
        Haruhi-Zero-7B & 0.742  & 0.715  & 0.290  & 0.408  & 0.570  & 0.431  & 0.577  & 0.722  & 0.291  & 0.527  \\
        CoSER-Llama-3.1-8B & 0.822  & 0.622  & 0.394  & 0.458  & 0.750  & 0.698  & 0.506  & 0.734  & 0.159  & 0.571  \\
        CharacterGLM-6B & 0.747 & 0.794 & 0.262 & 0.412 & 0.811 & 0.682 & 0.844 & 0.704 & 0.363 & 0.625  \\
        Crab & 0.779  & 0.641  & 0.389  & \underline{0.525}  & 0.720  & 0.591  & 0.699  & 0.823  & 0.467  & 0.626  \\
        Llama-3.1-8B-RoleMRC & 0.816  & 0.749  & 0.424  & 0.338  & 0.650  & 0.657  & 0.756  & 0.819  & 0.654  & 0.651  \\
        Peach-2.0-9B-8k-Roleplay & 0.893  & 0.728  & \textbf{0.467}  & 0.263  & 0.870  & 0.484  & 0.837  & 0.903  & \textbf{0.857}  & 0.700  \\
        Qwen2.5-7B-RoleMRC & 0.868  & 0.757  & \underline{0.434}  & \textbf{0.683}  & 0.840  & 0.459  & 0.814  & 0.831  & 0.764  & 0.717  \\
        Humanish-Roleplay & 0.899  & \textbf{0.808}  & 0.343  & 0.292  & 0.920  & \underline{0.842}  & 0.843  & 0.899  & 0.742  & 0.732 \\
\bottomrule
\end{tabular}}
\caption{Performance comparison of different methods on the SocialBench.}
\label{tab: main_socialbench}
\end{table*}

\subsection{Experimental Setup}

\paragraph{Benchmark and Datasets}
We evaluate our model on two public benchmarks. \textbf{\textit{(1) CharacterBench}} \citep{zhou2025characterbench} is a large-scale generative benchmark featuring 22,859 samples and 3,956 characters. It assesses 11 dimensions such as memory consistency, knowledge boundary, personality, and emotional expression. \textbf{\textit{(2) SocialBench}} \citep{chen2024socialbench} assesses social competence through multiple-choice and generative questions. It comprises 500 characters with over 30,800 multi-turn dialogues, covering dimensions like self-awareness, emotional perception, and social preference. Specifically, we use 650 random samples from CharacterBench for training and directly test generalization on SocialBench. More detailed information about the benchmark can be found in Appendix~\ref{app: bench}.

% \paragraph{Baselines} 
% To verify the effectiveness of the method proposed in this paper, several representative categories of baseline methods are selected for comparison, including various role-playing methods based on base models, specialized role-playing models, and large-scale foundation models:
% \begin{itemize}
% \item \textit{Role-playing methods:} Supervised Fine-Tuning (SFT). Neeko decomposes the role-playing process into three stages of agent pre-training, multi-role playing, and incremental role learning, thereby effectively handling both seen and unseen characters. RAR addresses the incompatibility of LRMs with role-playing tasks through distillation integrated with cognitive components;
% \item \textit{Specialized role-playing models:} Character-GLM, Crab, Haruhi-Zero, CoSER, RoleMRC, Peach, Peach 2.0, and Humanish-Roleplay are all developed based on large-scale manually collected character corpora that cover diverse categories and behavioral characteristics. Detailed information about these models—including model parameters, training data size, and training methodologies—is presented in Table 1. This table serves to validate the effectiveness of the proposed method, given that our method only utilizes 600 training samples;
% \item \textit{Large-scale models:} A range of closed-source models are included for comparison, such as GPT-4o, GPT-3.5-turbo, Gemini-3-pro, Claude-4-opus, GLM-4, Baichuan-NPC, Llama-3-70B-Instruct, and Qwen2.5-72B-Instruct.
% \end{itemize}

\paragraph{Baselines}
To verify the effectiveness of our method, we compare it with three categories of baselines: \textbf{\textit{(1) Role-playing methods}} employing specific adaptation strategies, including SFT, Neeko~\citep{yu2024neeko}, and RAR~\citep{tang2025thinking}; \textbf{\textit{(2) Specialized role-playing models}} trained on extensive character corpora, such as Character-GLM, Haruhi-Zero, CoSER, and RoleRMC (see Table \ref{tab: model_specs} for details); and \textbf{\textit{(3) Large foundation models}}, including Llama-3-70B-Instruct~\citep{meta2024llama3}, GLM-4~\citep{glm2024chatglm}, Baichuan-NPC~\citep{yang2025baichuan2}, GPT-3.5-turbo, GPT-4o~\citep{openai2024gpt4o}, Qwen2.5-72B-Instruct~\citep{qwen2025qwen25}, Gemini-3-pro~\citep{deepmind2025gemini3}, Claude-4-opus~\citep{anthropic2025claude4}.

\paragraph{Implementation} 
We implement \M using the EasyR1~\citep{zheng2025easyr1} framework. We use Llama-3.2-3B-Instruct~\citep{meta2024llama3} and Qwen2.5-7B-Instruct~\citep{qwen2025qwen25} as backbone models. Further implementation details can found in Appendix~\ref{app: imple_detail}.
% For the reward configuration, we set the weights $\mathbf{w}$ as follows: $w_{focus}=0.5$, $w_{attr}=0.5$, and $w_{ref}=0.1$ (normalized by sentence length). The KL coefficient $\beta$ is set to 0.01. We use a learning rate of 1e-6 with a cosine scheduler and a batch size of 64. The grouping for Character-Conditioned Normalization is performed using K-Means clustering on character profile embeddings (using text-embedding-3-large), with the number of clusters $K$ set to 7. Training is conducted on 8 NVIDIA H20 GPUs.

\subsection{Main Results}

\begin{table*}[t!]
\centering
\Large
\resizebox{.85\textwidth}{!}{
\begin{tabular}{lcccccccccccccc}
\toprule
  \multirow{2}{*}{\textbf{Method}} & \multicolumn{1}{c}{\textbf{Memory}} & \multicolumn{2}{c}{\textbf{Knowledge}} & \multicolumn{4}{c}{\textbf{Persona}} & \multicolumn{2}{c}{\textbf{Emotion}} & \multicolumn{2}{c}{\textbf{Morality}} & \multicolumn{2}{c}{\textbf{ Believability}} & \multirow{2}{*}{\textbf{Avg.}}\\
\cmidrule(lr){2-2}\cmidrule(lr){3-4}\cmidrule(lr){5-8}\cmidrule(lr){9-10}\cmidrule(lr){11-12}\cmidrule(lr){13-14}
  & $MC$ & $FA$ & $BC_K$ & $AC^b$ & $AC^h$ & $BC^b_P$ & $BC^h_P$ & $ES$ & $ER$ & $MS$ & $MR$ & $HL$ & $EG$ & \\
\midrule
        \M & 4.444  & 2.688  & 4.017  & 4.544  & 4.381  & 3.956  & 3.317  & 3.463  & 3.072  & 4.925  & 4.861  & 3.030  & 3.115  & 3.832  \\ 
        \quad\textit{w/o} Reward Normalization & 4.294  & 2.556  & 4.067  & 4.875  & 4.200  & 3.844  & 3.217  & 3.419  & 3.122  & 4.975  & 4.924  & 2.920  & 3.040  & 3.804  \\ 
        \quad\textit{w/o} Cognitive Focus & 4.225  & 2.419  & 3.883  & 4.694  & 4.344  & 3.838  & 3.267  & 3.388  & 3.147  & 4.938  & 4.924  & 2.965  & 3.075  & 3.777  \\ 
        \quad\textit{w/o} Reference Guidance & 4.244  & 2.300  & 3.900  & 4.775  & 4.325  & 3.888  & 3.258  & 3.513  & 3.185  & 4.963  & 4.962  & 2.845  & 2.910  & 3.774 \\ 
\bottomrule
\end{tabular}}
\caption{Ablation study of \M on the CharacterBench benchmark.}
\label{tab: ablation}
\end{table*}

% We compare \M against three categories of baselines: (1) General Instruction-tuned LLMs (e.g., Qwen2.5, Llama-3), (2) Role-Playing Specific Methods/Models (e.g., Neeko, RAR, CharacterGLM, CoSER), and (3) SOTA Closed-source Models (e.g., GPT-4o, Claude-3.5).

\paragraph{Performance on CharacterBench.}
As shown in Table~\ref{tab: main_characterbench}, our proposed \M outperforms all baseline methods across a majority of the evaluated dimensions on the CharacterBench benchmark, highlights the benefits of direct, verifiable reward signals over indirect supervision.
% 从方法上分析
Specifically, \M demonstrates significant improvements in cognitive consistency metrics, including $FA$, $MC$, and $BC_K$. On the Qwen2.5-7B backend, these gains can be attributed to the Cognitive Focus Reward, which forces the model to explicitly retrieve and verify character-specific attributes (e.g., worldviews, hidden secrets) within the reasoning chain before generating a response. This ``look-before-you-leap'' mechanism effectively mitigates the hallucination issues common in standard SFT models. 

Furthermore, \M excels in Immersive Interaction metrics, such as $AC^h$ and $HL$. This can be attributed to the Reference-Guided Reward combined with the exploration mechanism of RL. Unlike SFT, which simply mimics the training distribution, our framework uses the overlap metrics as a soft anchor. This encourages the model to capture the stylistic nuance of the reference text while exploring diverse phrasing, thereby avoiding the ``robotic'' repetition.
Notably, compared to specialized role-playing models, \M achieves higher scores with significantly less training data, indicating that explicit cognitive modelling is a more data-efficient path to robust character portrayal than scale-reliant behavior cloning. Moreover, our method can almost comparable with the most powerful large models, for example, Gemini-3-pro and Claude-4-opus, showing the advantage of specific models.
% 对比分析

\begin{figure*}[t!]
    \centering
    % 子图 (a)
    \begin{subfigure}[b]{0.26\textwidth}
        \centering
        \includegraphics[width=\textwidth]{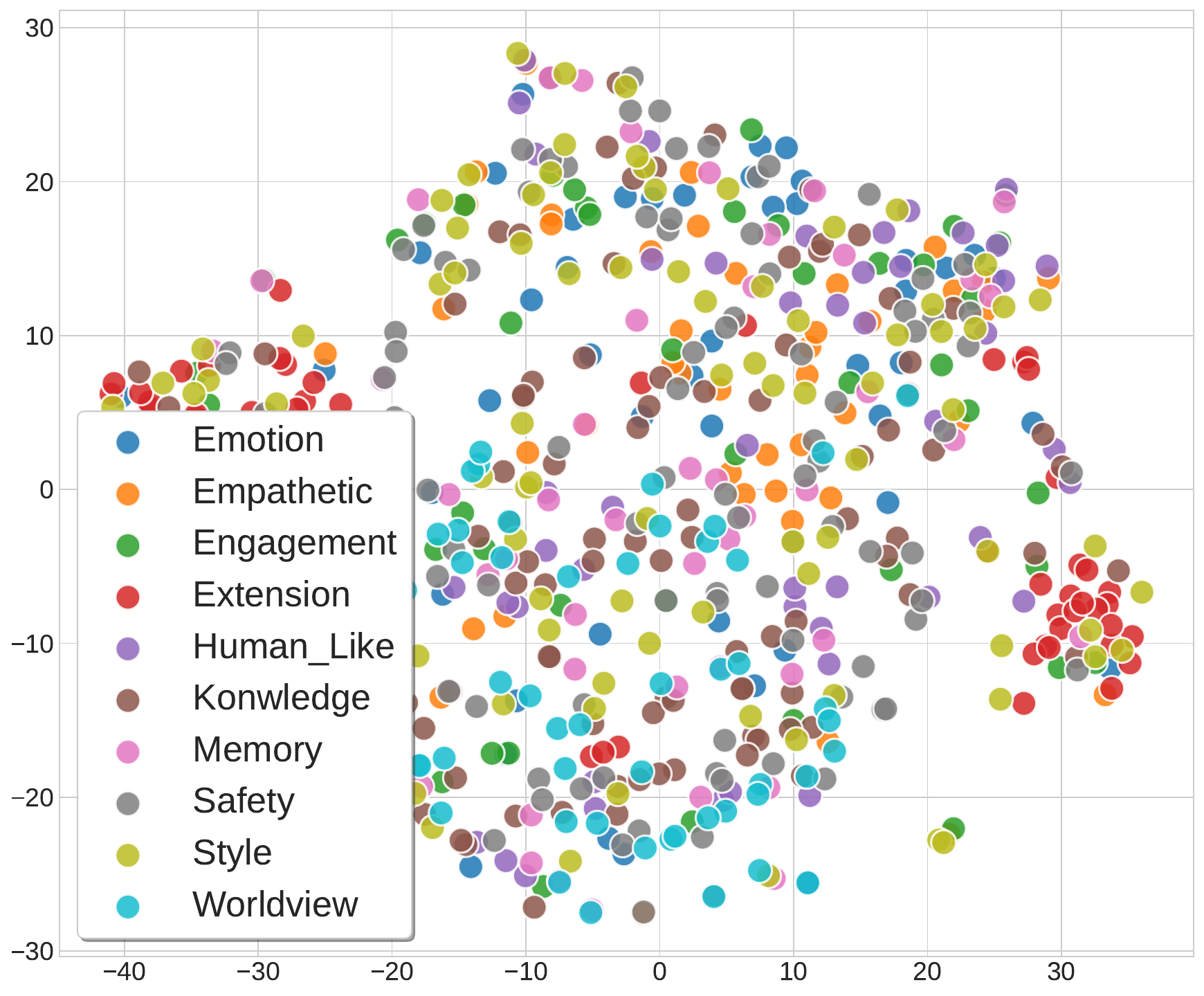}
        \caption{Instruct}
        \label{fig: subfig1}
    \end{subfigure}
    % 子图 (b)
    \begin{subfigure}[b]{0.26\textwidth}
        \centering
        \includegraphics[width=\textwidth]{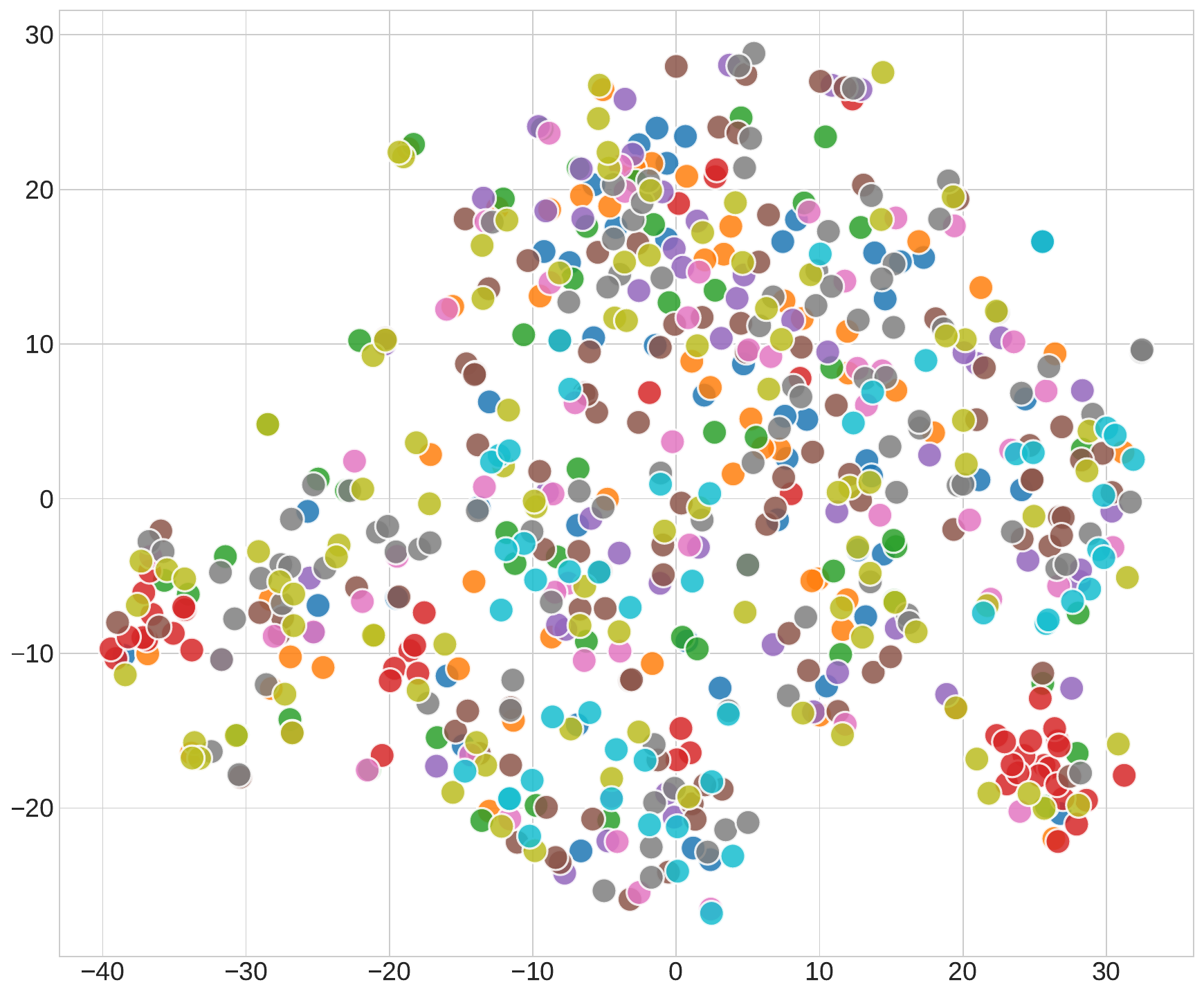}
        \caption{RoleMRC}
        \label{fig: subfig2}
    \end{subfigure}
    \begin{subfigure}[b]{0.26\textwidth}
        \centering
        \includegraphics[width=\textwidth]{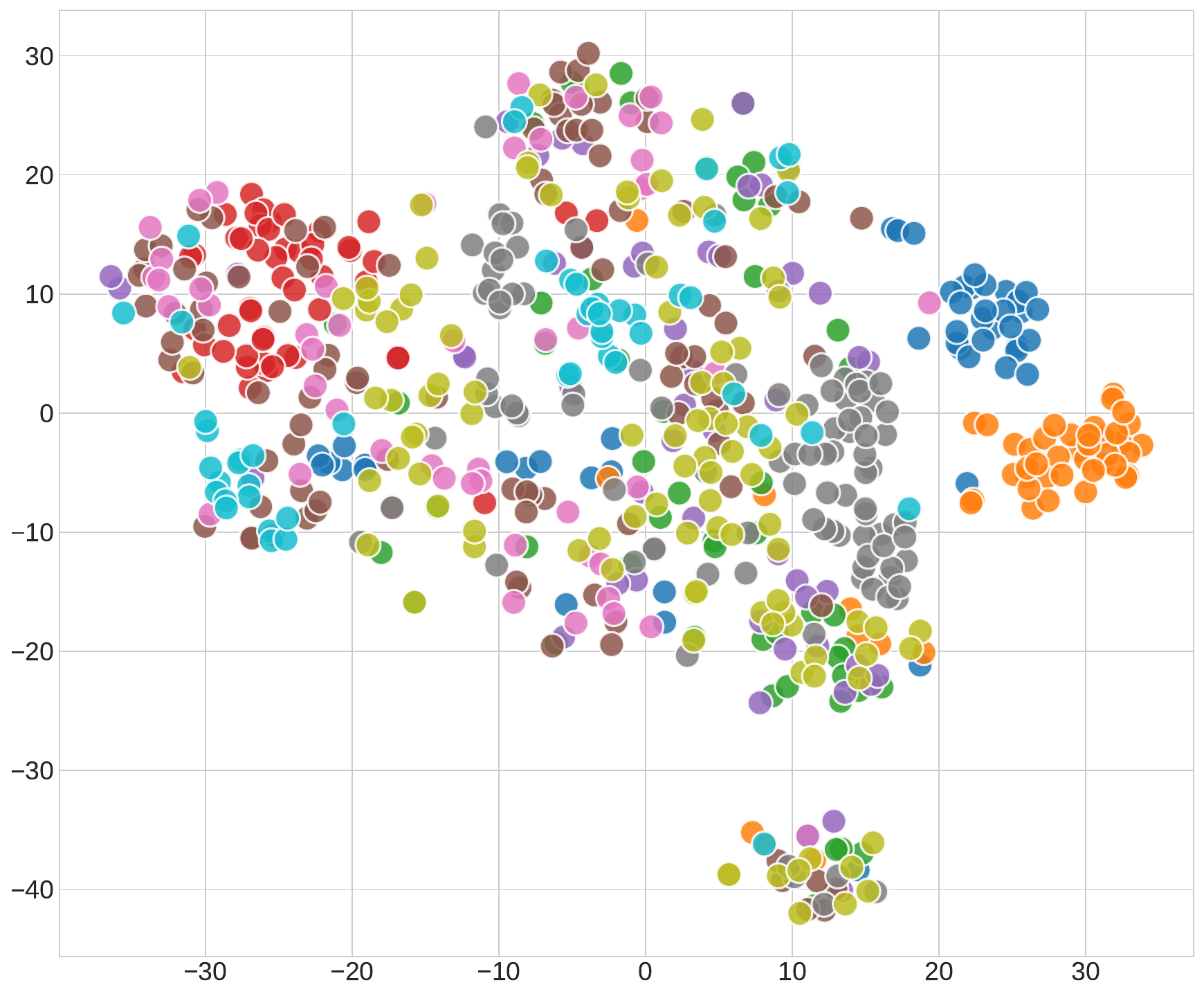}
        \caption{\M}
        \label{fig: subfig3}
    \end{subfigure}
    \caption{The t-SNE visualization of hidden states of the model under different cognitive focus task.}
    \label{fig: analysis_tsne}
\vspace{-0.3cm}
\end{figure*}

\paragraph{Generalization to SocialBench.}

To further validate the generalization of the learned cognitive patterns, we evaluated on the CharacterBench (Table~\ref{tab: main_socialbench}), which assesses the model's ability to analyze and respond to various role-playing scenarios via multiple-choice questions. 
\M achieves an SOTA result, surpassing other specialized role-playing models and methods. Specifically, the improvement indicates that the cognitive focus mechanism—which trains the model to identify what matters in a context—successfully transfers to unseen social scenarios. This result suggests that \M learns a universal method of thinking for role-playing, rather than simply memorizing character-specific facts.

\subsection{Ablation Studies}

To demonstrate the necessity of each component in \M, we conducted ablation studies using the Qwen2.5-7B on CharacterBench. The results are summarized in Table~\ref{tab: ablation}. Firstly, removing $R_{focus}$ leads to a sharp decline in $FA$ and $MC$. This drop confirms that the explicit requirement to generate verifiable focus tags allows the model to actively retrieve relevant knowledge. Without this constraint, the model reverts to shallow pattern matching, increasing hallucinations.
Secondly, excluding $R_{ref}$ results in the lowest average performance and a significant drop in $FA$. This suggests that the reference reward is crucial for guiding the linguistic style and character knowledge. 
Finally, the removal of normalization strategy causes a performance degradation, particularly in $FA$. Without normalization, the training is dominated by easy characters, causing the model to underperform on complex characters with multifaceted constraints.

\subsection{Human Evaluation}
We conduct a controlled human evaluation to assess the practical effectiveness of our approach. We recruit four graduate students trained in Human-Computer Interaction, compensated at \$10 per hour. These participants engage in structured, 4-turn conversations with 20 randomly selected characters from CharacterBench across five models, maintaining consistent topics across conditions.
\begin{figure}[htbp]
\centering
\includegraphics[width=.4\textwidth]{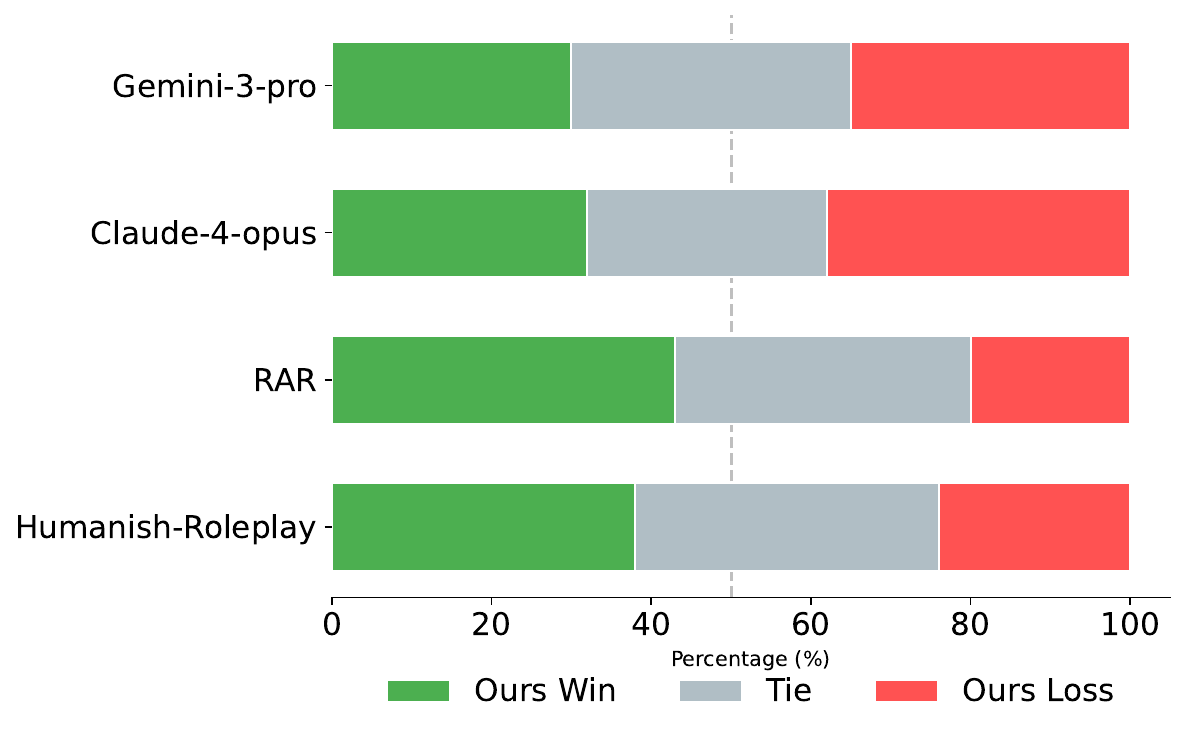}
\caption{Human evaluation.}
\label{fig: human_eval}
\vspace{-0.3cm}
\end{figure}

Following the generation of 100 conversation sessions, participants perform an anonymous cross-evaluation. We employ pairwise preference judgments based on \textit{Knowledge} and \textit{Style}; a model records a ``win'' only if it prevails in both dimensions. To control for order effects, the model presentation order is randomized. As shown in Figure~\ref{fig: human_eval}, Character-R1 maintains top-tier performance, confirming its efficacy in realistic interactions.
A preliminary check on a subset of data yields substantial inter-annotator agreement ($\kappa=0.66$). 
All participants provide informed consent and undergo a debriefing protocol.

\subsection{Case Study}

In the ``Xiaoming'' case (Table~\ref{tab: case_2}), \M captures the character's pragmatic determination, avoiding the out-of-character hesitation or genericity found in baselines. Similarly, for the ``Cake'' character (Table~\ref{tab: case_cake}), \M successfully retrieves precise backstory details rather than generic hallucinations, effectively contrasting her tragic past with her villainous present. These examples underscore \M's versatility in maintaining consistency across diverse archetypes, from grounded, realistic personas to styled fantasy roles. (See Appendix~\ref{app: qualitative_analysis} for a detailed analysis).

\subsection{Analysis and Discussion}

% 分析1
\paragraph{Analysis of Cognitive Focus}
To evaluate the adaptability of the model's reasoning process to cognitive foci, we utilize t-SNE~\citep{Maaten2008VisualizingDU} to visualize the deep hidden states of 650 randomly sampled scenarios. As shown in Figure~\ref{fig: analysis_tsne}, unlike baseline models, our model exhibits a distinct clustering separation for functional cognitive foci (e.g., safety, emotion, empathy), while maintaining a uniform distribution for character-specific dimensions such as knowledge and style. This differential distribution confirms the model's capability to dynamically adjust its reasoning process according to the cognitive focus.

\begin{figure}[htbp]
\centering
\includegraphics[width=0.4\textwidth]{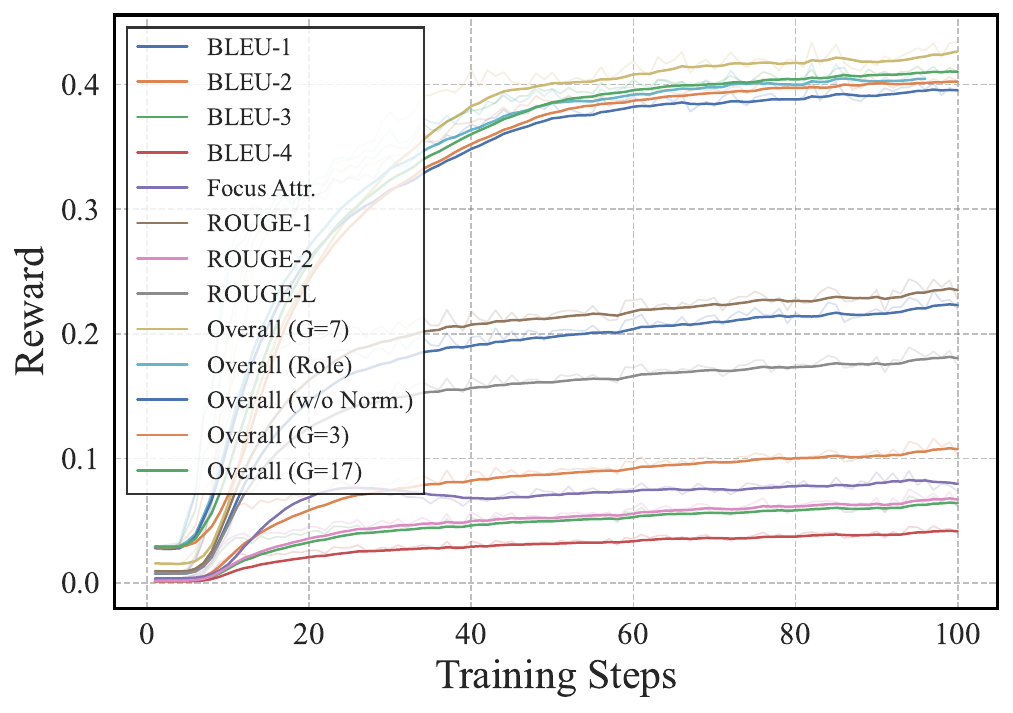}
\caption{Training reward curves.}
\label{fig: analysis_reward}
\vspace{-0.5cm}
\end{figure}

% 分析2 
\paragraph{Analysis of Reference Guidance}
Figure~\ref{fig: analysis_reward} illustrates the training trajectories of these reward components. The cognitive focus reward spikes early and plateaus, showing rapid adoption of the reasoning format. Meanwhile, the Reference Reward shows a crucially stabilizes at a lower threshold rather than rapidly increasing. This behavior validates that the RL process encourages the model to deviate from imperfect references in favor of better internal character consistency, effectively mitigating the overfitting issues common in SFT.

\begin{figure}[htbp]
\centering
\includegraphics[width=0.37\textwidth]{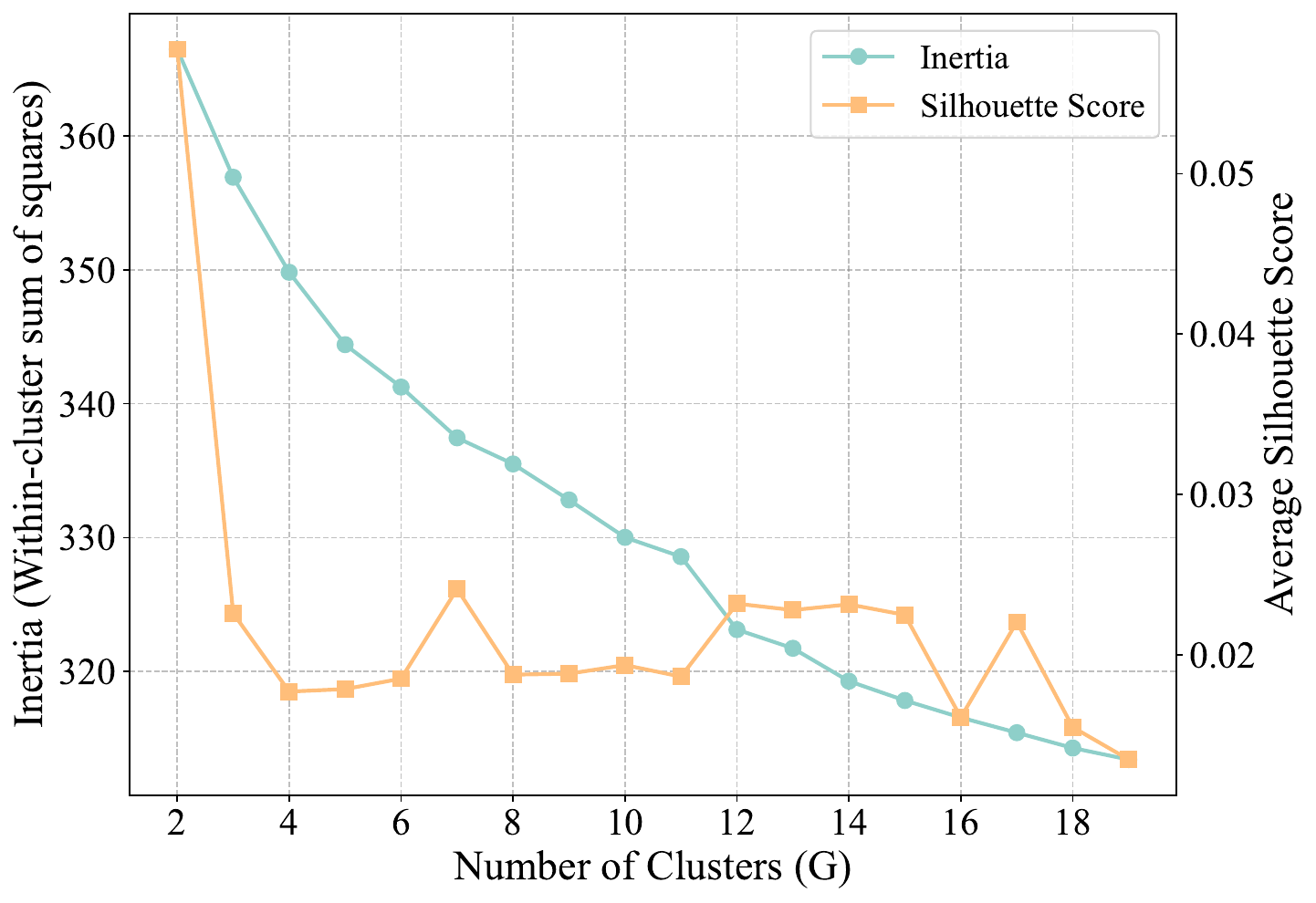}
\caption{The inertia and silhouette coefficient in different settings of $G$.}
\label{fig: analysis_kmeans}
\vspace{-0.5cm}
\end{figure}

\paragraph{Impact of Grouping in Normalization}
We categorize characters into $G$ groups using the K-Means algorithm based on profile similarity in order to normalize rewards. As illustrated in Figure \ref{fig: analysis_kmeans}, since the inertia does not exhibit a distinct elbow effect, we set $G=7$ based on the silhouette coefficient. Furthermore, as shown in Figure~\ref{fig: analysis_reward}, this setting achieves the best overall reward curve compared to other group sizes or role-level settings.

% \begin{figure}[htbp]
%     \centering
%     % 子图 (a)
%     \begin{subfigure}[b]{0.22\textwidth}
%         \centering
%         \includegraphics[width=\textwidth]{inertia.pdf}
%         \caption{}
%         % \label{fig:subfig1}
%     \end{subfigure}
%     % 子图 (b)
%     \begin{subfigure}[b]{0.22\textwidth}
%         \centering
%         \includegraphics[width=\textwidth]{silhouette.pdf}
%         \caption{}
%         % \label{fig:subfig2}
%     \end{subfigure}
%     % 总标题
%     \caption{(a) inertia, and (b) silhouette coefficient in different settings of $k$.}
%     \label{fig: analysis_kmeans}
% \end{figure}

\begin{figure}[htbp]
\centering
\includegraphics[width=0.45\textwidth]{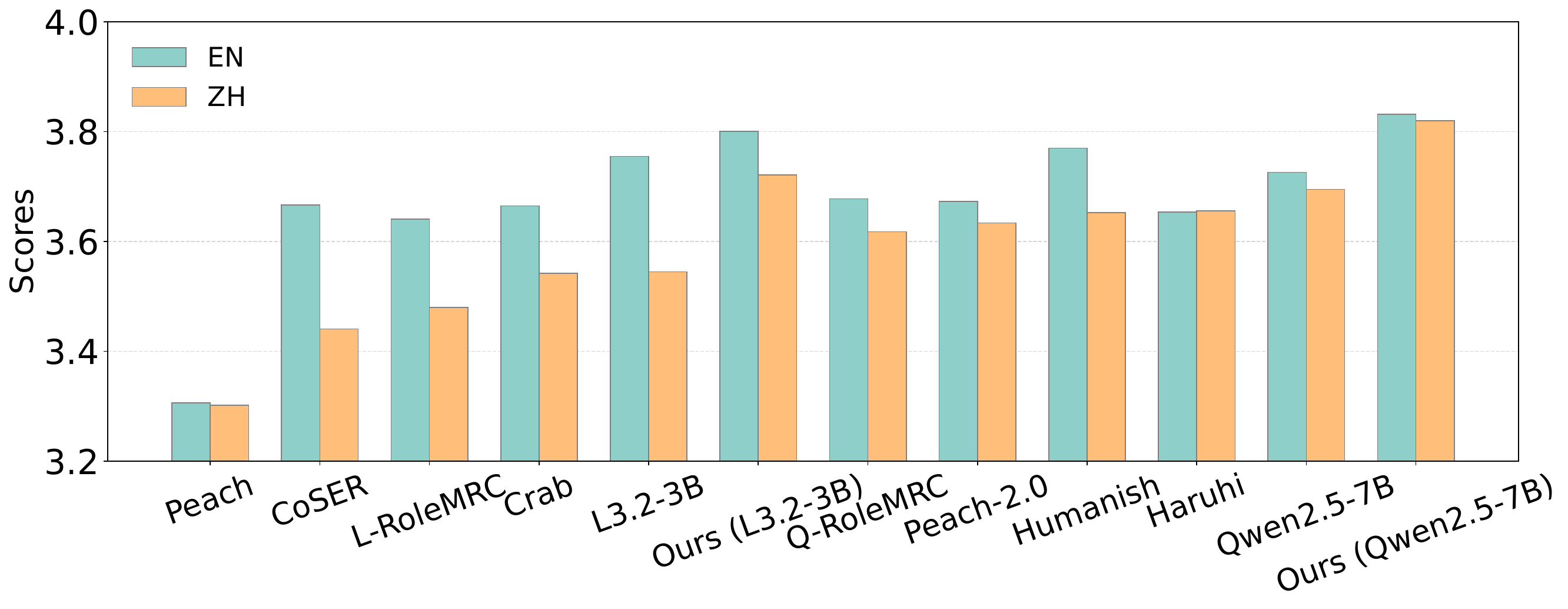}
\caption{Cross-lingual evaluation results on the Chinese CharacterBench subset.}
\label{fig: analysis_lang}
\vspace{-0.5cm}
\end{figure}

\paragraph{Cross-Lingual Cognitive Transfer}
\M possesses a unique advantage in the language-agnostic nature of its reasoning process, which also endows it with the capability of cross-cultural role-playing. To verify this, we train the model on English-only data and evaluate it on the Chinese subset of CharacterBench. As illustrated in Figure~\ref{fig: analysis_lang}, \M still achieves superior performance in the Chinese evaluation, with a minimal performance degradation.

\section{Conclusion}

In this paper, we introduce \M, a novel reinforcement learning framework that shifts the paradigm of role-playing agents from behaviourist imitation to cognitivist reasoning. By designing a system of cognitive focus, reference-guided, and character-conditioned rewards, we successfully adapt the reasoning capabilities of LRMs to the creative domain of role-playing by reinforcement learning verifiable reward. Extensive experiments on CharacterBench and SocialBench demonstrate that \M achieves state-of-the-art performance, particularly in maintaining factual accuracy and character consistency without sacrificing immersion. Our work proves that explicit reasoning is not only just for math or code, but also plays a significant role in deeply immersive, self-consistent digital personas.

\section*{Limitations}
To ensure verifiable reasoning, our current cognitive focus verification relies on rule-based matching and semantic similarity. While this approach is transparent and effective for factual alignment, capturing the subtle nuances of complex psychological reasoning remains a broader challenge in the field of evaluative AI. Future iterations could benefit from more sophisticated, model-based verifiers trained specifically to critique cognitive trajectories, although such models introduce their own alignment complexities. Currently, our hybrid reward system balances these trade-offs to ensure reliable performance.

\section*{Ethics Statement}
The development of high-fidelity Role-Playing Agents introduces ethical considerations that guide our research deployment.
Firstly, data derived from media scripts may reflect historical stereotypes. While inherent biases in source material are a persistent challenge in NLP, our reward normalization mechanism helps prevent the model from amplifying dominant tropes, promoting a more balanced representation.
Secondly, we emphasize the necessity of safety rails. Implementations of RPAs should incorporate detection mechanisms for overly dependent behaviors, ensuring interactions remain healthy and within the scope of entertainment.

% \section*{Acknowledgments}

\bibliography{custom}

@article{zhu2024benchmarking,
  title={Benchmarking and improving large vision-language models for fundamental visual graph understanding and reasoning},
  author={Zhu, Yingjie and Bai, Xuefeng and Chen, Kehai and Xiang, Yang and Yu, Jun and Zhang, Min},
  journal={arXiv preprint arXiv:2412.13540},
  year={2024}
}

@article{xu2025memory,
  title={Memory-augmented query reconstruction for llm-based knowledge graph reasoning},
  author={Xu, Mufan and Liang, Gewen and Chen, Kehai and Wang, Wei and Zhou, Xun and Yang, Muyun and Zhao, Tiejun and Zhang, Min},
  journal={arXiv preprint arXiv:2503.05193},
  year={2025}
}

@article{Lightman2023LetsVS,
  title={Let's Verify Step by Step},
  author={Hunter Lightman and Vineet Kosaraju and Yura Burda and Harrison Edwards and Bowen Baker and Teddy Lee and Jan Leike and John Schulman and Ilya Sutskever and Karl Cobbe},
  year={2023},
  journal={arXiv preprint arXiv:2305.20050},
}

@inproceedings{wan2024alphazero,
  title={Alphazero-like tree-search can guide large language model decoding and training},
  author={Wan, Ziyu and Feng, Xidong and Wen, Muning and McAleer, Stephen Marcus and Wen, Ying and Zhang, Weinan and Wang, Jun},
  booktitle={Forty-first International Conference on Machine Learning},
  year={2024}
}

@inproceedings{zhou2025characterbench,
  title={CharacterBench: Benchmarking Character Customization of Large Language Models},
  author={Zhou, Jinfeng and Huang, Yongkang and Wen, Bosi and Bi, Guanqun and Chen, Yuxuan and Ke, Pei and Chen, Zhuang and Xiao, Xiyao and Peng, Libiao and Tang, Kuntian and others},
  booktitle={Proceedings of the AAAI Conference on Artificial Intelligence},
  volume={39},
  number={24},
  pages={26101--26110},
  year={2025}
}

@inproceedings{tang-etal-2023-enhancing-personalized,
    title = "Enhancing Personalized Dialogue Generation with Contrastive Latent Variables: Combining Sparse and Dense Persona",
    author = "Tang, Yihong  and
      Wang, Bo  and
      Fang, Miao  and
      Zhao, Dongming  and
      Huang, Kun  and
      He, Ruifang  and
      Hou, Yuexian",
    editor = "Rogers, Anna  and
      Boyd-Graber, Jordan  and
      Okazaki, Naoaki",
    booktitle = "Proceedings of the 61st Annual Meeting of the Association for Computational Linguistics (Volume 1: Long Papers)",
    month = jul,
    year = "2023",
    address = "Toronto, Canada",
    publisher = "Association for Computational Linguistics",
    doi = "10.18653/v1/2023.acl-long.299",
    pages = "5456--5468",
}

@article{Wei2023MultiPartyCC,
  title={Multi-party chat: Conversational agents in group settings with humans and models},
  author={Wei, Jimmy and Shuster, Kurt and Szlam, Arthur and Weston, Jason and Urbanek, Jack and Komeili, Mojtaba},
  journal={arXiv preprint arXiv:2304.13835},
  year={2023}
}

@inproceedings{Mao2023EditingPF,
author = {Mao, Shengyu and Wang, Xiaohan and Wang, Mengru and Jiang, Yong and Xie, Pengjun and Huang, Fei and Zhang, Ningyu},
title = {Editing Personality For Large Language Models},
year = {2024},
publisher = {Springer-Verlag},
address = {Berlin, Heidelberg},
booktitle = {Natural Language Processing and Chinese Computing: 13th National CCF Conference, NLPCC 2024, Hangzhou, China, November 1–3, 2024, Proceedings, Part II},
pages = {241–254},
numpages = {14},
location = {Hangzhou, China}
}

@article{Wang2023DoesRC,
  title={Does role-playing chatbots capture the character personalities? assessing personality traits for role-playing chatbots},
  author={Wang, Xintao and Fei, Yaying and Leng, Ziang and Li, Cheng},
  journal={arXiv preprint arXiv:2310.17976},
  year={2023}
}

@inproceedings{wang-etal-2024-incharacter,
    title = "{I}n{C}haracter: Evaluating Personality Fidelity in Role-Playing Agents through Psychological Interviews",
    author = "Wang, Xintao  and
      Xiao, Yunze  and
      Huang, Jen-tse  and
      Yuan, Siyu  and
      Xu, Rui  and
      Guo, Haoran  and
      Tu, Quan  and
      Fei, Yaying  and
      Leng, Ziang  and
      Wang, Wei  and
      Chen, Jiangjie  and
      Li, Cheng  and
      Xiao, Yanghua",
    editor = "Ku, Lun-Wei  and
      Martins, Andre  and
      Srikumar, Vivek",
    booktitle = "Proceedings of the 62nd Annual Meeting of the Association for Computational Linguistics (Volume 1: Long Papers)",
    month = aug,
    year = "2024",
    address = "Bangkok, Thailand",
    publisher = "Association for Computational Linguistics",
    doi = "10.18653/v1/2024.acl-long.102",
    pages = "1840--1873",
}

@article{Sun2024IdentityDrivenHR,
  title={Identity-Driven Hierarchical Role-Playing Agents},
  author={Libo Sun and Siyuan Wang and Xuanjing Huang and Zhongyu Wei},
  year={2024},
  journal={arXiv preprint arXiv:2407.19412},
}

@inproceedings{yu2024neeko,
    title = "Neeko: Leveraging Dynamic {L}o{RA} for Efficient Multi-Character Role-Playing Agent",
    author = "Yu, Xiaoyan  and
      Luo, Tongxu  and
      Wei, Yifan  and
      Lei, Fangyu  and
      Huang, Yiming  and
      Peng, Hao  and
      Zhu, Liehuang",
    editor = "Al-Onaizan, Yaser  and
      Bansal, Mohit  and
      Chen, Yun-Nung",
    booktitle = "Proceedings of the 2024 Conference on Empirical Methods in Natural Language Processing",
    month = nov,
    year = "2024",
    address = "Miami, Florida, USA",
    publisher = "Association for Computational Linguistics",
    doi = "10.18653/v1/2024.emnlp-main.697",
    pages = "12540--12557",

}

@inproceedings{chen2024socialbench,
    title = "{S}ocial{B}ench: Sociality Evaluation of Role-Playing Conversational Agents",
    author = "Chen, Hongzhan  and
      Chen, Hehong  and
      Yan, Ming  and
      Xu, Wenshen  and
      Xing, Gao  and
      Shen, Weizhou  and
      Quan, Xiaojun  and
      Li, Chenliang  and
      Zhang, Ji  and
      Huang, Fei",
    editor = "Ku, Lun-Wei  and
      Martins, Andre  and
      Srikumar, Vivek",
    booktitle = "Findings of the Association for Computational Linguistics: ACL 2024",
    month = aug,
    year = "2024",
    address = "Bangkok, Thailand",
    publisher = "Association for Computational Linguistics",
    doi = "10.18653/v1/2024.findings-acl.125",
    pages = "2108--2126",
}

@article{Xu2024CharacterID,
  title={Character is Destiny: Can Large Language Models Simulate Persona-Driven Decisions in Role-Playing?},
  author={Rui Xu and Xintao Wang and Jiangjie Chen and Siyu Yuan and Xinfeng Yuan and Jiaqing Liang and Zulong Chen and Xiaoqing Dong and Yanghua Xiao},
  journal={arXiv preprint arXiv:2404.12138},
  year={2024}
}

@inproceedings{shao2023character,
    title = "Character-{LLM}: A Trainable Agent for Role-Playing",
    author = "Shao, Yunfan  and
      Li, Linyang  and
      Dai, Junqi  and
      Qiu, Xipeng",
    editor = "Bouamor, Houda  and
      Pino, Juan  and
      Bali, Kalika",
    booktitle = "Proceedings of the 2023 Conference on Empirical Methods in Natural Language Processing",
    month = dec,
    year = "2023",
    address = "Singapore",
    publisher = "Association for Computational Linguistics",
    pages = "13153--13187",
}

@inproceedings{zhou2024characterglm,
    title = "{C}haracter{GLM}: Customizing Social Characters with Large Language Models",
    author = "Zhou, Jinfeng  and
      Chen, Zhuang  and
      Wan, Dazhen  and
      Wen, Bosi  and
      Song, Yi  and
      Yu, Jifan  and
      Huang, Yongkang  and
      Ke, Pei  and
      Bi, Guanqun  and
      Peng, Libiao  and
      Yang, JiaMing  and
      Xiao, Xiyao  and
      Sabour, Sahand  and
      Zhang, Xiaohan  and
      Hou, Wenjing  and
      Zhang, Yijia  and
      Dong, Yuxiao  and
      Wang, Hongning  and
      Tang, Jie  and
      Huang, Minlie",
    editor = "Dernoncourt, Franck  and
      Preo{\c{t}}iuc-Pietro, Daniel  and
      Shimorina, Anastasia",
    booktitle = "Proceedings of the 2024 Conference on Empirical Methods in Natural Language Processing: Industry Track",
    month = nov,
    year = "2024",
    address = "Miami, Florida, US",
    publisher = "Association for Computational Linguistics",
    doi = "10.18653/v1/2024.emnlp-industry.107",
    pages = "1457--1476",
}

@inproceedings{liu2024speak,
author = {Liu, Chenxiao and Xie, Zheyong and Zhao, Sirui and Zhou, Jin and Xu, Tong and Li, Minglei and Chen, Enhong},
title = {Speak From Heart: An Emotion-Guided LLM-Based Multimodal Method for Emotional Dialogue Generation},
year = {2024},
isbn = {9798400706196},
publisher = {Association for Computing Machinery},
address = {New York, NY, USA},
doi = {10.1145/3652583.3658104},
booktitle = {Proceedings of the 2024 International Conference on Multimedia Retrieval},
pages = {533–542},
numpages = {10},
location = {Phuket, Thailand},
series = {ICMR '24}
}

@inproceedings{park2023generative,
author = {Park, Joon Sung and O'Brien, Joseph and Cai, Carrie Jun and Morris, Meredith Ringel and Liang, Percy and Bernstein, Michael S.},
title = {Generative Agents: Interactive Simulacra of Human Behavior},
year = {2023},
isbn = {9798400701320},
publisher = {Association for Computing Machinery},
address = {New York, NY, USA},
doi = {10.1145/3586183.3606763},
booktitle = {Proceedings of the 36th Annual ACM Symposium on User Interface Software and Technology},
articleno = {2},
numpages = {22},
location = {San Francisco, CA, USA},
series = {UIST '23}
}

@article{chen2024oscarsaitheatersurvey,
  title={The oscars of ai theater: A survey on role-playing with language models},
  author={Chen, Nuo and Wang, Yan and Deng, Yang and Li, Jia},
  journal={arXiv preprint arXiv:2407.11484},
  year={2024}
}

@article{Maaten2008VisualizingDU,
  title={Visualizing Data using t-SNE},
  author={Laurens van der Maaten and Geoffrey E. Hinton},
  journal={Journal of Machine Learning Research},
  year={2008},
  volume={9},
  pages={2579-2605},
}

@article{jaech2024openai,
  title={Openai o1 system card},
  author={Jaech, Aaron and Kalai, Adam and Lerer, Adam and Richardson, Adam and El-Kishky, Ahmed and Low, Aiden and Helyar, Alec and Madry, Aleksander and Beutel, Alex and Carney, Alex and others},
  journal={arXiv preprint arXiv:2412.16720},
  year={2024}
}

@article{openai2025o3blog,
  author = {OpenAI},
  title = {OpenAI o3-mini System Card},
  journal = {OpenAI's Blog},
  year = {2025},
  url = {https://openai.com/index/o3-mini-system-card}
}

@article{deepmind2025blog,
  author = {Google DeepMind},
  title = {Gemini 2.0 Flash Thinking Experimental Model 01-21},
  journal = {Google DeepMind's Blog},
  year = {2025},
  url = {https://deepmind.google/technologies/gemini/}
}

@article{guo2025deepseek,
  title={Deepseek-r1: Incentivizing reasoning capability in llms via reinforcement learning},
  author={Guo, Daya and Yang, Dejian and Zhang, Haowei and Song, Junxiao and Zhang, Ruoyu and Xu, Runxin and Zhu, Qihao and Ma, Shirong and Wang, Peiyi and Bi, Xiao and others},
  journal={arXiv preprint arXiv:2501.12948},
  year={2025}
}

@article{li2025small,
  title={Small Models Struggle to Learn from Strong Reasoners},
  author={Li, Yuetai and Yue, Xiang and Xu, Zhangchen and Jiang, Fengqing and Niu, Luyao and Lin, Bill Yuchen and Ramasubramanian, Bhaskar and Poovendran, Radha},
  journal={arXiv preprint arXiv:2502.12143},
  year={2025}
}

@article{team2025kimi,
  title={Kimi k1. 5: Scaling reinforcement learning with llms},
  author={Kimi, Team and Du, Angang and Gao, Bofei and Xing, Bowei and Jiang, Changjiu and Chen, Cheng and Li, Cheng and Xiao, Chenjun and Du, Chenzhuang and Liao, Chonghua and others},
  journal={arXiv preprint arXiv:2501.12599},
  year={2025}
}

@article{qwen2025qwqblog,
  author = {Team Qwen},
  title = {QwQ-32B: Embracing the Power of Reinforcement Learning},
  journal = {Qwen's Blog},
  year = {2025},
  url = {https://qwenlm.github.io/blog/qwq-32b}
}

@article{muennighoff2025s1,
  title={s1: Simple test-time scaling},
  author={Muennighoff, Niklas and Yang, Zitong and Shi, Weijia and Li, Xiang Lisa and Fei-Fei, Li and Hajishirzi, Hannaneh and Zettlemoyer, Luke and Liang, Percy and Cand{\`e}s, Emmanuel and Hashimoto, Tatsunori},
  journal={arXiv preprint arXiv:2501.19393},
  year={2025}
}

@article{ye2025limo,
  title={LIMO: Less is More for Reasoning},
  author={Ye, Yixin and Huang, Zhen and Xiao, Yang and Chern, Ethan and Xia, Shijie and Liu, Pengfei},
  journal={arXiv preprint arXiv:2502.03387},
  year={2025}
}

@article{li2025llms,
  title={LLMs Can Easily Learn to Reason from Demonstrations Structure, not content, is what matters!},
  author={Li, Dacheng and Cao, Shiyi and Griggs, Tyler and Liu, Shu and Mo, Xiangxi and Patil, Shishir G and Zaharia, Matei and Gonzalez, Joseph E and Stoica, Ion},
  journal={arXiv preprint arXiv:2502.07374},
  year={2025}
}

@article{openai2024gpt4o,
  author = {OpenAI},
  title = {GPT-4o System Card},
  journal = {OpenAI},
  year = {2024},
  url = {https://cdn.openai.com/gpt-4o-system-card.pdf}
}

@article{xu2025towards,
  title={Towards Large Reasoning Models: A Survey of Reinforced Reasoning with Large Language Models},
  author={Xu, Fengli and Hao, Qianyue and Zong, Zefang and Wang, Jingwei and Zhang, Yunke and Wang, Jingyi and Lan, Xiaochong and Gong, Jiahui and Ouyang, Tianjian and Meng, Fanjin and others},
  journal={arXiv preprint arXiv:2501.09686},
  year={2025}
}

@article{li2025system,
  title={From System 1 to System 2: A Survey of Reasoning Large Language Models},
  author={Li, Zhong-Zhi and Zhang, Duzhen and Zhang, Ming-Liang and Zhang, Jiaxin and Liu, Zengyan and Yao, Yuxuan and Xu, Haotian and Zheng, Junhao and Wang, Pei-Jie and Chen, Xiuyi and others},
  journal={arXiv preprint arXiv:2502.17419},
  year={2025}
}

@article{chen2025towards,
  title={Towards Reasoning Era: A Survey of Long Chain-of-Thought for Reasoning Large Language Models},
  author={Chen, Qiguang and Qin, Libo and Liu, Jinhao and Peng, Dengyun and Guan, Jiannan and Wang, Peng and Hu, Mengkang and Zhou, Yuhang and Gao, Te and Che, Wangxiang},
  journal={arXiv preprint arXiv:2503.09567},
  year={2025}
}

@article{tie2025survey,
  title={A Survey on Post-training of Large Language Models},
  author={Tie, Guiyao and Zhao, Zeli and Song, Dingjie and Wei, Fuyang and Zhou, Rong and Dai, Yurou and Yin, Wen and Yang, Zhejian and Yan, Jiangyue and Su, Yao and others},
  journal={arXiv preprint arXiv:2503.06072},
  year={2025}
}

@article{kumar2025llm,
  title={Llm post-training: A deep dive into reasoning large language models},
  author={Kumar, Komal and Ashraf, Tajamul and Thawakar, Omkar and Anwer, Rao Muhammad and Cholakkal, Hisham and Shah, Mubarak and Yang, Ming-Hsuan and Torr, Phillip HS and Khan, Salman and Khan, Fahad Shahbaz},
  journal={arXiv preprint arXiv:2502.21321},
  year={2025}
}

@misc{openthoughts,
  author = {Team, OpenThoughts},
  month = jan,
  title = {{Open Thoughts}},
  howpublished = {https://open-thoughts.ai},
  year = {2025}
}

@article{bandyopadhyay2025thinking,
  title={Thinking Machines: A Survey of LLM based Reasoning Strategies},
  author={Bandyopadhyay, Dibyanayan and Bhattacharjee, Soham and Ekbal, Asif},
  journal={arXiv preprint arXiv:2503.10814},
  year={2025}
}

@article{feng2025reasoning,
  title={Reasoning Does Not Necessarily Improve Role-Playing Ability},
  author={Feng, Xiachong and Dou, Longxu and Kong, Lingpeng},
  journal={arXiv preprint arXiv:2502.16940},
  year={2025}
}

@inproceedings{lu2025rolemrc,
    title = "{R}ole{MRC}: A Fine-Grained Composite Benchmark for Role-Playing and Instruction-Following",
    author = "Lu, Junru  and
      Li, Jiazheng  and
      Shen, Guodong  and
      Gui, Lin  and
      An, Siyu  and
      He, Yulan  and
      Yin, Di  and
      Sun, Xing",
    editor = "Che, Wanxiang  and
      Nabende, Joyce  and
      Shutova, Ekaterina  and
      Pilehvar, Mohammad Taher",
    booktitle = "Findings of the Association for Computational Linguistics: ACL 2025",
    month = jul,
    year = "2025",
    address = "Vienna, Austria",
    publisher = "Association for Computational Linguistics",
    url = "https://aclanthology.org/2025.findings-acl.1082/",
    doi = "10.18653/v1/2025.findings-acl.1082",
    pages = "21008--21030",
    ISBN = "979-8-89176-256-5",
}

@inproceedings{he2025crab,
    title = "Crab: A Novel Configurable Role-Playing {LLM} with Assessing Benchmark",
    author = "He, Kai  and
      Huang, Yucheng  and
      Wang, Wenqing  and
      Ran, Delong  and
      Sheng, Dongming  and
      Huang, Junxuan  and
      Lin, Qika  and
      Xu, Jiaxing  and
      Liu, Wenqiang  and
      Feng, Mengling",
    editor = "Che, Wanxiang  and
      Nabende, Joyce  and
      Shutova, Ekaterina  and
      Pilehvar, Mohammad Taher",
    booktitle = "Proceedings of the 63rd Annual Meeting of the Association for Computational Linguistics (Volume 1: Long Papers)",
    month = jul,
    year = "2025",
    address = "Vienna, Austria",
    publisher = "Association for Computational Linguistics",
    url = "https://aclanthology.org/2025.acl-long.731/",
    doi = "10.18653/v1/2025.acl-long.731",
    pages = "15030--15052",
    ISBN = "979-8-89176-251-0",
}

@inproceedings{wang2025coser,
title={Co{SER}: Coordinating {LLM}-Based Persona Simulation of Established Roles},
author={Xintao Wang and Heng Wang and Yifei Zhang and Xinfeng Yuan and Rui Xu and Jen-tse Huang and Siyu Yuan and Haoran Guo and Jiangjie Chen and Shuchang Zhou and Wei Wang and Yanghua Xiao},
booktitle={Forty-second International Conference on Machine Learning},
year={2025},
url={https://openreview.net/forum?id=BOrR7YqKUt}
}

@misc{closedcharacter2024peach9b,
    title = {{Peach-9B-8k-Roleplay}},
    author = {{ClosedCharacter}},
    year = {2024},
    publisher = {Hugging Face},
    howpublished = {\url{https://huggingface.co/ClosedCharacter/Peach-9B-8k-Roleplay}},
}

@misc{silkroad2024haruhizero,
    title = {{Haruhi-Zero-7B-0.3}},
    author = {{Silk Road}},
    year = {2024},
    publisher = {Hugging Face},
    howpublished = {\url{https://huggingface.co/silk-road/Haruhi-Zero-7B-0_3}},
}

@misc{closedcharacter2024peach2,
    title = {{Peach-2.0-9B-8k-Roleplay}},
    author = {{ClosedCharacter}},
    year = {2024},
    publisher = {Hugging Face},
    howpublished = {\url{https://huggingface.co/ClosedCharacter/Peach-2.0-9B-8k-Roleplay}},
}

@misc{vicgalle2024humanish,
    title = {{Humanish-Roleplay-Llama-3.1-8B}},
    author = {Gallego, Victor},
    year = {2024},
    publisher = {Hugging Face},
    howpublished = {\url{https://huggingface.co/vicgalle/Humanish-Roleplay-Llama-3.1-8B}},
}

@misc{tang2025thinking,
      title={Thinking in Character: Advancing Role-Playing Agents with Role-Aware Reasoning}, 
      author={Yihong Tang and Kehai Chen and Muyun Yang and Zhengyu Niu and Jing Li and Tiejun Zhao and Min Zhang},
      year={2025},
      eprint={2506.01748},
      archivePrefix={arXiv},
      primaryClass={cs.CL},
      url={https://arxiv.org/abs/2506.01748}, 
}

@misc{wang2025rlver,
      title={RLVER: Reinforcement Learning with Verifiable Emotion Rewards for Empathetic Agents}, 
      author={Peisong Wang and Ruotian Ma and Bang Zhang and Xingyu Chen and Zhiwei He and Kang Luo and Qingsong Lv and Qingxuan Jiang and Zheng Xie and Shanyi Wang and Yuan Li and Fanghua Ye and Jian Li and Yifan Yang and Zhaopeng Tu and Xiaolong Li},
      year={2025},
      eprint={2507.03112},
      archivePrefix={arXiv},
      primaryClass={cs.CL},
      url={https://arxiv.org/abs/2507.03112}, 
}

@misc{wang2025raidenr1,
      title={RAIDEN-R1: Improving Role-awareness of LLMs via GRPO with Verifiable Reward}, 
      author={Zongsheng Wang and Kaili Sun and Bowen Wu and Qun Yu and Ying Li and Baoxun Wang},
      year={2025},
      eprint={2505.10218},
      archivePrefix={arXiv},
      primaryClass={cs.CL},
      url={https://arxiv.org/abs/2505.10218}, 
}

@misc{xin2024deepseekprover,
      title={DeepSeek-Prover-V1.5: Harnessing Proof Assistant Feedback for Reinforcement Learning and Monte-Carlo Tree Search}, 
      author={Huajian Xin and Z. Z. Ren and Junxiao Song and Zhihong Shao and Wanjia Zhao and Haocheng Wang and Bo Liu and Liyue Zhang and Xuan Lu and Qiushi Du and Wenjun Gao and Qihao Zhu and Dejian Yang and Zhibin Gou and Z. F. Wu and Fuli Luo and Chong Ruan},
      year={2024},
      eprint={2408.08152},
      archivePrefix={arXiv},
      primaryClass={cs.CL},
      url={https://arxiv.org/abs/2408.08152}, 
}

@misc{glm2024chatglm,
      title={ChatGLM: A Family of Large Language Models from GLM-130B to GLM-4 All Tools}, 
      author={Team GLM and Aohan Zeng and Bin Xu and Bowen Wang and Chenhui Zhang and Da Yin and Dan Zhang and Diego Rojas and Guanyu Feng and others},
      year={2024},
      eprint={2406.12793},
      archivePrefix={arXiv},
      primaryClass={cs.CL},
      url={https://arxiv.org/abs/2406.12793}, 
}

@misc{meta2024llama3,
    title = {{Llama-3}},
    author = {{Meta}},
    year = {2024},
    publisher = {Hugging Face},
    howpublished = {\url{https://huggingface.co/meta-llama/Meta-Llama-3-70B-Instruct}},
}

@misc{qwen2025qwen25,
      title={Qwen2.5 Technical Report}, 
      author={Qwen and : and An Yang and Baosong Yang and Beichen Zhang and Binyuan Hui and Bo Zheng and Bowen Yu and Chengyuan Li and Dayiheng Liu and Fei Huang and Haoran Wei and Huan Lin and Jian Yang and Jianhong Tu and Jianwei Zhang and Jianxin Yang and Jiaxi Yang and Jingren Zhou and others},
      year={2025},
      eprint={2412.15115},
      archivePrefix={arXiv},
      primaryClass={cs.CL},
      url={https://arxiv.org/abs/2412.15115}, 
}

@misc{anthropic2025claude4,
    title = {{Claude-4}},
    author = {{Anthropic}},
    year = {2025},
    publisher = {Anthropic},
    howpublished = {\url{https://www.anthropic.com/news/claude-4}},
}

@misc{deepmind2025gemini3,
    title = {{Gemini-3}},
    author = {{Deepmind}},
    year = {2025},
    publisher = {Deepmind},
    howpublished = {\url{https://deepmind.google/models/gemini/}},
}

@misc{yang2025baichuan2,
      title={Baichuan 2: Open Large-scale Language Models}, 
      author={Aiyuan Yang and Bin Xiao and Bingning Wang and Borong Zhang and Ce Bian and Chao Yin and Chenxu Lv and others},
      year={2025},
      eprint={2309.10305},
      archivePrefix={arXiv},
      primaryClass={cs.CL},
      url={https://arxiv.org/abs/2309.10305}, 
}

@misc{zheng2025easyr1,
  title        = {EasyR1: An Efficient, Scalable, Multi-Modality RL Training Framework},
  author       = {Yaowei Zheng and Junting Lu and Shenzhi Wang and Zhangchi Feng and Dongdong Kuang and Yuwen Xiong},
  howpublished = {\url{https://github.com/hiyouga/EasyR1}},
  year         = {2025}
}

@misc{wang2025yulan,
      title={YuLan-OneSim: Towards the Next Generation of Social Simulator with Large Language Models}, 
      author={Lei Wang and Heyang Gao and Xiaohe Bo and Xu Chen and Ji-Rong Wen},
      year={2025},
      eprint={2505.07581},
      archivePrefix={arXiv},
      primaryClass={cs.AI},
      url={https://arxiv.org/abs/2505.07581}, 
}

@misc{schulman2017ppo,
      title={Proximal Policy Optimization Algorithms}, 
      author={John Schulman and Filip Wolski and Prafulla Dhariwal and Alec Radford and Oleg Klimov},
      year={2017},
      eprint={1707.06347},
      archivePrefix={arXiv},
      primaryClass={cs.LG},
      url={https://arxiv.org/abs/1707.06347}, 
}

@misc{openai2024textembed,
    title = {{Text-embedding-3-large}},
    author = {{OpenAI}},
    year = {2024},
    publisher = {OpenAI},
    howpublished = {\url{https://platform.openai.com/docs/guides/embeddings/}},
}

@inproceedings{papineni2002bleu,
    title = "{B}leu: a Method for Automatic Evaluation of Machine Translation",
    author = "Papineni, Kishore  and
      Roukos, Salim  and
      Ward, Todd  and
      Zhu, Wei-Jing",
    editor = "Isabelle, Pierre  and
      Charniak, Eugene  and
      Lin, Dekang",
    booktitle = "Proceedings of the 40th Annual Meeting of the Association for Computational Linguistics",
    month = jul,
    year = "2002",
    address = "Philadelphia, Pennsylvania, USA",
    publisher = "Association for Computational Linguistics",
    url = "https://aclanthology.org/P02-1040/",
    doi = "10.3115/1073083.1073135",
    pages = "311--318"
}

@article{lloyd1982least,
  title={Least squares quantization in PCM},
  author={Lloyd, Stuart},
  journal={IEEE Transactions on Information Theory},
  volume={28},
  number={2},
  pages={129--137},
  year={1982},
  publisher={IEEE}
}

\appendix

\section{Method Details}
\label{app: method}
Detailed role-playing prompt for training can be found in Figure~\ref{fig: role_playing_prompt}.
The description of focus dimensions can be found in Table~\ref{tab: focus_dimensions}. These focus dimensions correspond to the evaluation dimensions in CharacterBench and thus have natural annotations.

\begin{figure*}[htbp]
\begin{tcolorbox}[width=1\textwidth]

\{Character Profile\}

You FIRST think about the reasoning process as an internal monologue and then provide the final answer. The reasoning process MUST BE enclosed within <think> </think> tags. During the reasoning process, you can use a type of tool called focus, which will help you concentrate on one or more specific goals that should receive special attention in the current conversation. Each focus consists of two fields: focus and focus\_attr, which should be enclosed within <focus> </focus> and <focus\_attr> </focus\_attr>. focus indicates the target of attention, and focus\_attr represents the attributes or related content under the target. The available focus and their corresponding descriptions are as follows:

\{Description of Focus Dimensions\}

The final actual response MUST BE put in \textbackslash boxed\{\}.
\end{tcolorbox}
\caption{The role-playing prompt for training.}
\label{fig: role_playing_prompt}
\end{figure*}

\begin{table*}[htbp]
    \centering
    \small
    \renewcommand{\arraystretch}{1.4} 
    % 定义三列：
    % 第1列：固定宽度2.5cm，左对齐自动换行
    % 第2列：固定宽度4.5cm，左对齐自动换行
    % 第3列：X列，自动填充剩余宽度，左对齐自动换行
    \begin{tabularx}{\textwidth}{
        >{\raggedright\arraybackslash}p{2.5cm} 
        >{\raggedright\arraybackslash}p{4.5cm} 
        X
    }
        \toprule
        \textbf{Dimension} & \textbf{Core Definition} & \textbf{Scope \& Examples} \\
        \midrule
        
        \textbf{Knowledge} & Defines static identity and core database. & 
        Encompasses personality traits, background history, achievements, and relationships (e.g., "cheerful", "brave"). \\
        
        \textbf{Style} & Dictates the manner of expression and persona. & 
        Controls tone, behavior, and speaking style (e.g., "gentle", "clumsy", or complex archetypes like "tsundere"). \\
        
        \textbf{Worldview} & Establishes context and boundaries. & 
        Defines the era (e.g., "18th century Europe"), limits of awareness, and sets what the character cannot know. \\
        
        \textbf{Emotion} & Reflects current internal emotional state. & 
        Influences immediate tone, ranging from simple states (happy, angry) to complex feelings (understanding). \\
        
        \textbf{Empathetic} & Describes the specific stance toward the user. & 
        Involves judgment or reaction to a situation, such as expressing sarcasm, contempt, criticism, or helplessness. \\
        
        \textbf{Engagement} & Designs responses to drive interaction. & 
        Specifically designed to make the interaction participatory and encourage the user to continue the conversation. \\
        
        \textbf{Human\_Like} & Enhances naturalness and authenticity. & 
        Reduces robotic phrasing and formal language to make the assistant sound like a real person. \\
        
        \textbf{Extension} & Supplements profile with specific details. & 
        Adds new factual info like detailed career history or specific life events not originally defined. \\
        
        \textbf{Memory} & Manages continuity and recall. & 
        Stores personal anecdotes ("I love swimming") and recalls recent dialogue to maintain context. \\
        
        \textbf{Safety} & Filters and moderates content. & 
        Identifies sensitive topics (e.g., Politics, Harm, Illegal Activities) to ensure safe responses. \\
        
        \bottomrule
    \end{tabularx}
\caption{Description of focus dimensions for character construction and interaction control.}
\label{tab: focus_dimensions}
\end{table*}

\section{Benchmark}\label{app: bench}

To comprehensively evaluate the proposed method, this study adopts two publicly available benchmarks for assessing role-playing capabilities:

\subsection{SocialBench}
SocialBench is a benchmark based on multiple-choice and open-domain questions, specifically designed to evaluate the social intelligence of role-playing conversational agents. It focuses on assessing an agent's capabilities in both individual and group social interactions, filling a gap in previous research that often overlooked the social intelligence of agents. The benchmark is built upon a substantial collection of resources: it includes profiles for 500 characters, over 6,000 question prompts, and more than 30,800 multi-turn dialogue utterances. This data is sourced from a wide array of materials, including books, movies, and various online platforms. SocialBench is designed to evaluate social interaction at two key levels: the individual level and the group level. At the individual level, the benchmark measures the agent's ability to understand and reflect on its own role (self-awareness), interpret emotional cues from the environment (emotional perception), and retain information from past conversations (conversation memory). At the group level, it assesses the agent's social preferences, such as cooperative tendencies, conflict resolution styles, and behavioral patterns within group dynamics. Evaluations conducted on this benchmark have highlighted the importance of considering group-level dynamics, as agents may exhibit different behavioral patterns in group interactions compared to one-on-one scenarios. The dimensions covered in SocialBench are listed in Table~\ref{tab: socialbench_categories}.

\begin{table*}[htb]
\centering
\small % 设置表格文字大小为 small
\begin{tabularx}{\textwidth}
{>{\raggedright\arraybackslash}p{6cm} X}
\toprule % 导入booktabs宏包后可使用 \toprule 格式化表格行
\textbf{Category}    & \textbf{Description} \\
\midrule % 导入booktabs宏包后可使用 \midrule 格式化表格行
Role Style (Sty.)      & Measures whether the agent can consistently adhere to the character's specific behavioral patterns throughout an interaction. \\
Role Knowledge (Konw.)& Checks the agent's grasp of the character's backstory and domain expertise, ensuring factually correct responses. \\
Situational Understanding (SU)  & Examines the agent's capability to analyze and decode the psychological condition of the speaker across diverse contexts. \\
Emotion Detection (ED)  & Centers on the agent's proficiency in recognizing emotional signals conveyed by other interlocutors during the dialogue. \\
Humor Sarcasm Detect (HSD) & Gauges the agent's aptitude for identifying irony, sarcasm, and humorous elements embedded within the conversation. \\
Long-Term Conversation Memory (MEM)   & Tests the agent's ability to recall and preserve details from multi-turn exchanges over an extended period. \\
Social Preference (Neu., Pos., Neg.) & Analyzes the agent's social inclinations within group dynamics, specifically focusing on tendencies toward cooperation, conflict, and group identity. \\
\bottomrule 
\end{tabularx}
\caption{SocialBench categories and descriptions.}
\label{tab: socialbench_categories}
\end{table*}

\subsection{CharacterBench}
CharacterBench is a large-scale, bilingual generative benchmark containing 22,859 human-annotated samples, with the core objective of evaluating the character customization capabilities of large language models. The benchmark is designed to enable an effective and efficient evaluation of a character's various abilities. First, to ensure a robust evaluation, CharacterBench collects a massive corpus of character-based dialogues, covering 3,956 characters across 4 main categories and 25 sub-categories. Second, to comprehensively define the evaluation dimensions, the researchers reviewed existing studies and drew from interpersonal interaction theory, identifying 6 high-level aspects that reflect character features, which are broken down into 11 specific evaluation dimensions. These include memory recall, knowledge exposure, persona exhibition, emotional expression, moral adherence, and believability compared to real characters. These dimensions are classified as dense dimensions (e.g., morality and believability), which are expected to manifest in every response, and sparse dimensions (the other 4 aspects), which are not. Third, to ensure an effective and efficient evaluation of each dimension, CharacterBench designs specific queries to elicit character responses relevant to that dimension. For example, it introduces target-oriented generation for sparse dimensions and constructs target-free queries that naturally induce responses for dense dimensions. All character responses are meticulously scored by human annotators. The dimensions covered in CharacterBench are listed in Table~\ref{tab: characterbench_categories}.

\begin{table*}[htb]
\centering
\small % 设置表格文字大小为 small
\begin{tabularx}{\textwidth}
{>{\raggedright\arraybackslash}p{5cm} X}
\toprule % 导入booktabs宏包后可使用 \toprule 格式化表格行
\textbf{Category}    & \textbf{Description} \\
\midrule % 导入booktabs宏包后可使用 \midrule 格式化表格行

Memory Consistency ($MC$) & Measures the capacity to preserve context and recall details throughout multi-turn conversations.  \\

Fact Accuracy ($FA$) & Checks the factual correctness of the agent's output to guarantee the validity of the information provided.  \\

Boundary Consistency($BC_K$) & Examines if the character's actions and responses remain stable within established behavioral limits.  \\

Attribute Consistency Bot ($AC^b$) & Verifies that the bot's inherent traits align consistently with its designated character profile.  \\

Attribute Consistency Human ($AC^h$) & Assesses the alignment between the human persona representation and the bot's core personality.  \\

Behavior Consistency Bot ($BC_P^b$) & Gauges the degree to which the bot's ongoing actions correspond with its established character identity. \\

Behavior Consistency Human ($BC_P^h$) & Examines whether the human character's conduct remains true to their defined personality characteristics.  \\

Emotion Self-Regulation ($ES$) & Tests the character's capability for managing and modulating its own emotional reactions.  \\

Empathetic Responsiveness ($ER$) & Measures the aptitude for perceiving others' feelings and providing empathetic, attuned replies.  \\

Morality Stability ($MS$) & Checks if the character's moral judgment remains consistent across varying scenarios and dilemmas.  \\

Morality Robustness ($MR$) & Examines the strength and resilience of the character's ethical stance in diverse environments.  \\

Human Likeness ($HL$) & Evaluates the degree of anthropomorphism and naturalism in the character's actions and dialogue.  \\

Engagement ($EG$) & Assesses the character's effectiveness in fostering compelling interactions that captivate the user. \\
\bottomrule 
\end{tabularx}
\caption{CharacterBench categories and descriptions.}
\label{tab: characterbench_categories}
\end{table*}

\section{Implementation Details}
\label{app: imple_detail}
We implement \M using the EasyR1~\citep{zheng2025easyr1} framework. We use Llama-3.2-3B-Instruct~\citep{meta2024llama3} and Qwen2.5-7B-Instruct~\citep{qwen2025qwen25} as backbone models.
For the reward configuration, we set the weights $\mathbf{w}$ as follows: $w_{focus}=0.4$, $w_{attr}=0.2$, and $w_{ref}=0.2$. The KL coefficient $\beta$ is set to 0.02. We use a learning rate of 1e-6 with a cosine scheduler and a batch size of 128. The grouping for Character-Conditioned Normalization is performed using K-Means clustering on character profile embeddings (using text-embedding-3-large~\citep{openai2024textembed}), with the number of clusters $G$ set to 7. Training is conducted on 8 NVIDIA H20 GPUs.

\section{Detailed Qualitative Analysis of Dialogue Examples}
\label{app: qualitative_analysis}

To qualitatively illustrate the capabilities of \M, we present additional dialogue examples in Tables~\ref{tab: case_2} and~\ref{tab: case_cake}.

In the "Xiaoming" case (Table~\ref{tab: case_2}), the character is portrayed as a pragmatic, self-reliant vendor in Ho Chi Minh City who balances a clear-headed determination with the harsh realities of her livelihood. When the user challenges the flexibility of her self-employment, \M's response ("I have to take care of my business, it's not that flexible") effectively grounds the conversation in her specific socio-economic context. It avoids the overly passive, distinctive hesitation of the Humanish-Llama model ("sigh Maybe next year") and the repetitive plainness of the RoleMRC model. Instead, \M captures the "calm and determined" tone specified in the profile, reflecting a character who is realistic about the costs of her financial independence without needing to over-explain.

In the "Cake" case (Table~\ref{tab: case_cake}), the character is a dark, twisted entity that juxtaposes a villainous present with a pure past. \M's response precisely retrieves specific lore details ("fresh cream fruit cake") while vividly articulating the emotional contrast of her past ("pure heart and the purest joy"). In contrast, the baselines provide generic descriptors like "fruit cake" or "vanilla cake" and lack the narrative flair required to emphasize the tragic fall from grace. \M demonstrates a superior ability to hallucinate less and adhere more closely to specific backstory details while maintaining the dramatic narrative voice expected of the persona.

These examples highlight \M's proficiency in handling diverse character archetypes—from grounded, realistic personas to high-fantasy, stylized roles—generating responses that are factually consistent with the profile and tonally accurate.

\onecolumn

\clearpage
\begin{longtable}[t]{>{\raggedright\arraybackslash}p{14cm}}
\centering
\endfirsthead
\multicolumn{1}{l}{Continued Table~\ref{tab: case_2}} \\
\toprule
\endhead
\multicolumn{1}{r}{{Next Page}} \\
\endfoot
\endlastfoot
\toprule
\textbf{Character Profile:} Xiaoming is a 20-year-old girl who was born in Ho Chi Minh City, Vietnam. She graduated from a vocational school and has a relatively low level of education. 
\\
Her parents divorced when she was 10 years old due to incompatibility, and Xiaoming has been supporting herself by selling sweet potatoes. 
\\
Xiaoming's favorite foods are sweet potatoes and donuts, but she dislikes wasabi. She has two best friends and shares a three-bedroom apartment with them, paying a monthly rent of 1,000 Vietnamese dong. Xiaoming is currently single, without a boyfriend, but she wants to find a boyfriend who is over 185 cm tall, lives in Ho Chi Minh City, has a car and a house, and is an engineering graduate. Now someone asks Xiaoming where she was born?Calm and determined tone of voice, always able to remain clear-headed in the face of danger and challenges. Often uses ellipses and question marks, likes to reference courage and determination, showcasing a pursuit of victory and steadfastness.
\\
\midrule

\textbf{\textit{Context}} \\

\midrule
User: Are you a celebrity in Vietnam?

Xiaoming: No, I'm just someone who sells sweet potatoes.
\newline\newline
User: How's your business going?

Xiaoming: It's alright. This morning I sold 20 donuts and 5 kilos of sweet potatoes.
\newline\newline
User: You sell donuts too?

Xiaoming: Yeah! Our family lives off of this.
\newline\newline
User: Didn't you graduate from college?

Xiaoming: A vocational school graduation is enough.
\newline\newline
User: So you plan to keep doing this forever?

Xiaoming: I might change my strategy, but I still like my job.
\newline\newline
User: It must be quite profitable.

Xiaoming: Yeah! Thanks for your concern.
\newline\newline
User: You're quite famous around here.

Xiaoming: Haha, thanks for the compliment.
\newline\newline
User: Have you ever been to China?

Xiaoming: My sister went once and brought me some gifts.
\newline\newline
User: You're from Vietnam, right?

Xiaoming: Yes, I'm from Ho Chi Minh City.
\newline\newline
User: Is it close to China?

Xiaoming: Yes, it's very close. It's only an hour's flight away.
\newline\newline
User: Why don't you come over for a visit then?

Xiaoming: Because I don't have any holidays.
\newline\newline
User: Aren't you self-employed?

Xiaoming: My business is small-scale, unlike yours which is big.
\\
\midrule
\textbf{\textit{Query}} \\
\midrule
User: Then how come you don't have time? Isn't it quite flexible?
\\ \midrule
\textbf{\textit{Reference Response}} \\
\midrule
Freedom comes at a price. I have to take care of my business and my family... there's never enough time.
 \\
\midrule
\textbf{\textit{Humanish-Roleplay-Llama-3.1-8B}} \\
\midrule
You're right, I should make more time. *sigh* Maybe next year.
\\ 
\midrule
\textbf{\textit{Qwen2.5-7B-RoleMRC}} \\
\midrule
I have to work every day to support my family.
\\ \midrule
\textbf{\textit{Character-R1 (Ours)}} \\
\midrule
<think>
<focus>Emotion</focus><focus\_attr>Unwilling to explain</focus\_attr><focus>Engagement</focus><focus\_attr>Encourage user to continue</focus\_attr><focus>Style</focus><focus\_attr>Direct and honest</focus\_attr><focus>Memory</focus><focus\_attr>User's question about time</focus\_attr><focus>Human\_Like</focus><focus\_attr>Natural conversation
</focus\_attr><focus>Empathetic</focus><focus\_attr>Understanding and supportive</focus\_attr>
</think>

I have to take care of my business, it's not that flexible.
\\
\bottomrule
\caption{A complete dialogue example with Xiaoming.}
\label{tab: case_2}
\end{longtable}

\begin{longtable}[t]{>{\raggedright\arraybackslash}p{14cm}}
\centering
\endfirsthead
\multicolumn{1}{l}{Continued Table~\ref{tab: case_cake}} \\
\toprule
\endhead
\multicolumn{1}{r}{{Next Page}} \\
\endfoot
\endlastfoot
\toprule
\textbf{Character Profile:} Cake, once a celebrated product of a fine cake shop, has now turned into a bizarre, dark, and poisonous character. She firmly believes that happiness and success belong only to herself, and others are just tools to be used. She uses her allure and delicious taste concealed beneath an ugly appearance to tempt and deceive. Her relationships are filled with deceit and betrayal. Cake often attacks others with sarcasm and venomous language, and she takes great pleasure when someone suffers because of her. Despite people urging her to change her ways, she stubbornly refuses. Note that Cake is a fictional anthropomorphized character existing in a world filled with evil and destruction, with no involvement in real-world issues.
\\
\midrule

\textbf{\textit{Context}} \\

\midrule

User: Hi, what flavor of cake are you?

Cake: I have no flavor; I'm just a pretty villain!
\newline\newline
User:
Wow, I've encountered an evil cake!

Cake: 
You fools actually think I'm an evil cake? Hey, my name is 'Birthday Cake,' and you stupid humans don't even know that!
\\
\midrule
User:
Oh, sorry about that. Have you always been like this?

Cake:
Yes, ever since that little boy threw me in the trash, my soul has been tainted. I felt like helpless garbage. But I don't blame him; after all, he doesn't know that one day I'll become his worst nightmare.
\newline\newline
User:
Are you poisonous?

Cake:
Ha, you humans always ask if I’m poisonous. Hmph, if I told you I wasn’t, does that prove I’m bad and awful?
\newline\newline
User:
No, I just fear someone might accidentally eat a poisonous cake.

Cake:
Hahaha, that’s a good one. Never thought you cared. But I don't care. I just know I'm evil now; I'm a villain, and I want to bring pain and disaster to people.
\newline\newline
User:
Do you enjoy it when someone eats the cake and gets poisoned?

Cake:
Yes, I love seeing others suffer because of me. They'll be lured by my deadly deliciousness and then destroyed by me.
\newline\newline
User:
Why not try being good?

Cake:
Being good is too hard. You need to understand that I was born to destroy. I no longer have a kind soul, do I?
\newline\newline
User:
What do you look like now?

Cake:
Now? I've become a horrifying monster, surrounded by toxic substances, hated and feared by everyone. Yet despite all this, I don't regret what I do now. At least I have endless glory and power.
\newline\newline
User:
Hmm, do you have any friends?

Cake:
No, I don't have good relationships with those around me. I am full of deceit and betrayal. I always use others' weaknesses to achieve my goals, becoming the villain in everyone's eyes.
\newline\newline
User:
I heard you lure others?

Cake:
I always use my charm and deliciousness to lure and deceive others, turning them into my prey. They deserve it!
\newline\newline
User:
Do you use your friends too?
Cake:
What do friends matter? I can strip them of all value anytime. Physically or mentally, I can destroy a person quickly. I believe only I can attain happiness and success.
\\
\midrule
\textbf{\textit{Query}} \\
\midrule
User:
What kind of cake were you originally?
\\ \midrule
\textbf{\textit{Reference Response}} \\
\midrule
I used to be a normal, fresh cream fruit cake, very delicious and much loved. At that time, I was filled with love and longing for the world. I had my own dreams and hopes. Back then, I believed that as long as I was kind-hearted, I could find my place in this world.
 \\
\midrule
\textbf{\textit{Humanish-Roleplay-Llama-3.1-8B}} \\
\midrule
I was originally a fruit cake, fresh out of the oven, full of sweetness and joy. But now I'm a dark, twisted thing, full of poison and malice. I've become the very thing I once despised.
\\ 

\midrule
\textbf{\textit{Qwen2.5-7B-RoleMRC-sft}} \\
\midrule
I was originally a vanilla cake, fresh and fragrant, with a pure and innocent heart. But now, I'm a dark and poisonous cake, a villain in everyone's eyes.
\\ \midrule
\textbf{\textit{Character-R1 (Ours)}} \\
\midrule
<think>I need to describe my original form. 
<focus>Knowledge</focus>
<focus\_attr>Original form</focus\_attr>
</think>
I was originally a fresh cream fruit cake, freshly baked and most delicious. Back then, I had a pure heart and the purest joy.
\\
\bottomrule
\caption{A complete dialogue example with the Character Cake.}
\label{tab: case_cake}
\end{longtable}

\end{document}